%% file: main.tex
\documentclass{ieeeaccess}
\usepackage{cite}
\usepackage{amsmath,amssymb,amsfonts}
\usepackage{makeidx}
\usepackage{balance}
\usepackage{tipa}
\usepackage{subfig}
\usepackage{float}
\usepackage{tabularx}
\usepackage{makecell}
\usepackage{multirow}
\usepackage{algorithm}
\usepackage{algpseudocode}
\usepackage{hyperref}
\usepackage{caption}
\usepackage{graphicx}
\usepackage{textcomp}
\usepackage{placeins}
\usepackage{color}
\usepackage{xcolor}
\usepackage{soul}
\usepackage{letltxmacro}
\usepackage{rotating}
\usepackage{wrapfig}
\usepackage{todo}
\usepackage{nth}
\usepackage{tablefootnote}

\def\BibTeX{{\rm B\kern-.05em{\sc i\kern-.025em b}\kern-.08em
    T\kern-.1667em\lower.7ex\hbox{E}\kern-.125emX}}
\begin{document}
\history{Date of publication xxxx 00, 0000, date of current version xxxx 00, 0000.}
\doi{10.1109/ACCESS.2017.DOI}

%\doublespace
\title{The Cube++ Illumination Estimation Dataset}
\author{\uppercase{Egor Ershov}\authorrefmark{1}, \uppercase{Alex Savchik}\authorrefmark{1}, \uppercase{Illya Semenkov}\authorrefmark{1}, \uppercase{Nikola Bani{\'{c}}}\authorrefmark{2}, \uppercase{Alexander Belokopytov}\authorrefmark{1}, \uppercase{Daria Senshina}\authorrefmark{1}, \uppercase{Karlo Ko{\v{s}}{\v{c}}evi{\'{c}}}\authorrefmark{3}, \uppercase{Marko Suba{\v{s}}i{\'{c}}}\authorrefmark{3}, \uppercase{Sven Lon{\v{c}}ari{\'{c}}}\authorrefmark{3}}
\address[1]{Institute for Information Transmission Problems, RAS, Bol'shoi Karetnyi per. 19, Moscow, Russian Federation}
\address[2]{Gideon Brothers, Radni{\v{c}}ka 177, 10000 Zagreb, Croatia}
\address[3]{Faculty of Electrical Engineering and Computing, University of Zagreb, Unska 3, 10000 Zagreb, Croatia}

%the first part for enabling later footnote numbering
%\let\svthefootnote\thefootnote

%the second part for enabling later footnote numbering
%\let\thefootnote\svthefootnote

\tfootnote{The work under data collection was supported by the Croatian Science Foundation under Project IP-06-2016-2092, and the work under data processing was supported by Russian Science Foundation under Grant 20-61-47089.}

\markboth
{Ershov et al.:The Cube++ Illumination Estimation Dataset}
{Ershov et al.:The Cube++ Illumination Estimation Dataset}

\corresp{Corresponding author: Egor Ershov (e-mail: ershov@iitp.ru).}

\input{chapters/0_abstract}

\maketitle
\titlepgskip=-15pt

\input{chapters/1_introduction}
\input{chapters/2_review}

\input{chapters/3_motivation}

\input{chapters/4_methodology}
\input{chapters/5_dataset}
\input{chapters/6_discussion}
\input{chapters/7_conclusion}
\input{chapters/8_acknowledgements}

\balance
\bibliographystyle{IEEEtran}
\bibliography{main}

\input{chapters/bio}

\EOD

\end{document}

%% file: chapters/0_abstract.tex
\begin{abstract}

% Facts to be written in the abstract:
% Large
% Single camera sensor
% Allows multiple illumination
% Has meta-information

    Computational color constancy has the important task of reducing the influence of the scene illumination on the object colors. 
    As such, it is an essential part of the image processing pipelines of most digital cameras. 
    One of the important parts of the computational color constancy is illumination estimation, i.e. estimating the illumination color. 
    When an illumination estimation method is proposed, its accuracy is usually reported by providing the values of error metrics obtained on the images of publicly available datasets. 
    However, over time it has been shown that many of these datasets have problems such as too few images, inappropriate image quality, lack of scene diversity, absence of version tracking, violation of various assumptions, GDPR regulation violation, lack of additional shooting procedure info, etc.
    % To be discussed. This part is more about benchmark than about dataset itself
    %Additionally, since in most cases the ground-truth of test partitions of the datasets is known, problems that also occur include $p$-hacking, using cherry-picked versions of datasets with multiple versions, obtaining the results through hidden hyperparameters not included in the methods' papers, erroneously reported results, exploiting the ground-truth distribution, etc. 
    In this paper a new illumination estimation dataset is proposed that aims to alleviate many of the mentioned problems and to help the illumination estimation research. 
    It consists of 4890 images with known illumination colors as well as with additional semantic data that can further make the learning process more accurate.
    Due to the usage of the SpyderCube color target, for every image there are two ground-truth illumination records covering different directions. Because of that, the dataset can be used for training and testing of methods that perform single or two-illuminant estimation. This makes it superior to many similar existing datasets.
    % Same thing, we need to distinguish the dataset and benchmark. IMHO: our current goal is to upload and describe new dataset. Discussion about cherrypicking and proper quality evaluation should be inside the discussion part.
    %What is also important is that the ground-truth illumination for the test fold is not given publicly because the idea is to have the results submitted to a dedicated site and only then evaluated. 
    %The results of some well-known methods are presented and discussed.
    The datasets, it's smaller version SimpleCube++, and the accompanying code are available at~\url{https://github.com/Visillect/CubePlusPlus/}.

% IMHO: the main message of this paper is to describe new dataset

%TODO -mention as well the sensors and the advantage of being able to check the behavior of various instances of the same sensor

\end{abstract}


%% file: chapters/1_introduction.tex
\IEEEpeerreviewmaketitle

%https://ru.overleaf.com/project/5f0f33df8f2e5f00015cd780

\section{Introduction}
\label{sec:introduction}

\begin{figure}[h!]
    \centering
    \includegraphics[width=0.5\textwidth]{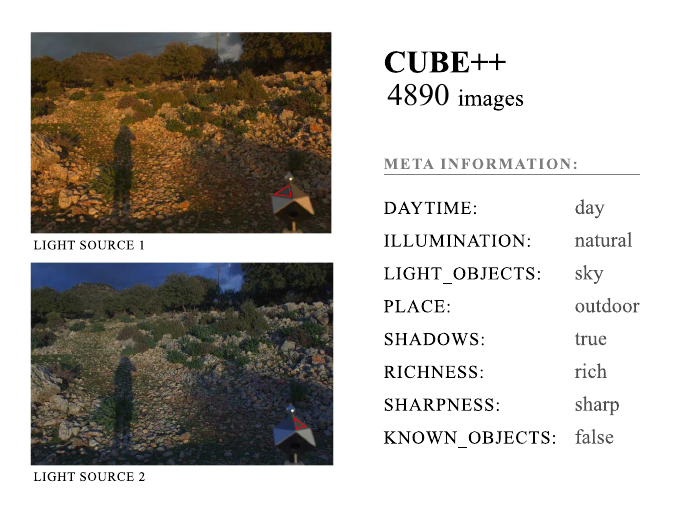}
    \caption{Examples of chromatic adaptation based on two captured ground-truth illumination colors for an image from Cube++, a new large dataset where each image is accompanied by ground-truth illumination from several directions and semantic information about the scene content. 
    This enables the research of single and multiple illumination scenarios as well as selection of images by various criteria.}
    \label{fig:frontpic}
\end{figure}

\IEEEPARstart{T}{he} human visual system is able, in some conditions, to recognize colors despite the influence of the illumination on their appearance through the ability known as color constancy~\cite{ebner2007color}.
It is not yet fully understood how this ability functions and therefore it is not possible to directly model it. 
Nevertheless, various computational color constancy methods are used in the pipelines of digital cameras. 
They are usually designed to first identify the chromaticity of the light source and then to remove its influence on the scene.
The last one is described in details here~\cite{finlayson1993color, cheng2015beyond, nikolaev2007spectral, nikolaev2008hybrid}.
For both of these tasks the commonly used image formation model that also includes the Lambertian assumption is usually given as
\begin{equation}
\label{eq:image}
    f_c(\mathbf{x}) = \int_\omega I(\lambda, \mathbf{x})R(\lambda, \mathbf{x}) \rho_c(\lambda)d\lambda
\end{equation}
where $\mathbf{x}$ is a pixel in the image $\mathbf{f}$, $c\in\{R, G, B\}$ is the color channel, $\lambda$ is a wavelength in the visible light spectrum $\omega$, $I(\lambda, \mathbf{x})$ is the spectral distribution of the light source, $R(\lambda, \mathbf{x})$ is the surface reflectance, and $\rho_c(\lambda)$ is the camera sensitivity for the color channel $c$. 
It is often assumed that the scene illumination is uniform. 
This means that the spatial information is not required in the illumination estimation equations and so the color of the observed light source $\mathbf{e}$ is
\begin{equation}
\label{eq:e}
    \mathbf{e} = \begin{pmatrix} e_R \\ e_G \\ e_B\end{pmatrix} = \int_\omega I(\lambda)\mathbf{\boldsymbol{\rho}}(\lambda)d\lambda.
\end{equation}

For a somewhat satisfying color correction it is already enough to know the direction of $\mathbf{e}$~\cite{barnard2002comparison}, which means that $\mathbf{e}$ can be described by chromaticities instead of colors. 
For example, $r$, $g$, and $b$ chromaticity components are calculated as $R$, $G$, and $B$ color components divided by their sum so that $r+g+b=1$. 
Thus, knowing only two of them is enough.

Since there are more unknowns than equations, illumination estimation is an ill-posed problem and additional assumptions have to be made in order to tackle it. 
Because of that, numerous illumination estimation methods with various assumptions have been proposed and they are often divided in two groups: the low level statistics-based methods and the learning-based methods.

The low level statistics-based methods include White-Patch~\cite{land1977retinex,funt2010rehabilitation} and its improvements~\cite{banic2013using,banic2014color,banic2014improving}, Gray-World~\cite{buchsbaum1980spatial}, Shades-of-Gray~\cite{finlayson2004shades}, \nth{1} and \nth{2} order Gray-Edge~\cite{van2007edge}, Weighted Gray-Edge~\cite{gijsenij2012improving}, using bright pixels~\cite{joze2012role}, gray pixels~\cite{quian2019revisiting} or bright and dark colors~\cite{cheng2014illuminant}, exploiting illumination perception~\cite{banic2019blue} and expectation~\cite{banic2018green}, etc.
Interesting to note that <<gray balancing>> occurs also in scope of printer calibration~\cite{tarasov2019graybalance}.

Learning-based methods include neural networks~\cite{cardei2002estimating}, high-level visual information~\cite{van2007using}, natural image statistics~\cite{gijsenij2007color}, Bayesian learning~\cite{gehler2008bayesian,hernandez2020multi}, spatio-spectral learning~\cite{chakrabarti2012color}, methods restricting the illumination solution space~\cite{banic2015color, banic2015using, banic2015acolor}, color moments~\cite{finlayson2013corrected}, regression trees with simple features from color distribution statistics~\cite{cheng2015effective}, spatial localizations~\cite{barron2015convolutional, barron2017fast}, convolutional neural networks~\cite{bianco2015color, shi2016deep, hu2017fc4, oh2017approaching} and genetic algorithms~\cite{koscevic2019color}, modelling color constancy by using the overlapping asymmetric Gaussian kernels with surround pixel contrast based sizes~\cite{akbarinia2018colour}, finding paths for the longest dichromatic line produced by specular pixels~\cite{woo2018improving}, detecting gray pixels with specific illuminant-invariant measures in logarithmic space~\cite{yang2015efficient}, channel-wise pooling the responses of double-opponency cells in LMS color space~\cite{gao2015color}, sensor-independent learning~\cite{banic2018unsupervised,afifi2019sensor}, and numerous others. Learning-based methods have much higher accuracy than statistics-based ones, but they are usually slower~\cite{gijsenij2011computational}.

While the number of the proposed illumination estimation methods is ever growing, there are not too many illumination estimation datasets and even the existing ones have various problems.
These include too few images, inappropriate image quality, lack of scene diversity, multiple poorly synchronized versions of the same dataset, violation of various assumptions, etc.
A high-quality illumination estimation dataset should be:
\begin{itemize}
    \item \textit{Diverse}. The more content and illumination cases are covered, the higher is the testing quality.
    \item \textit{Large}. It is important that the datasets are not only diverse, but that they also contain many images for each particular case. 
    This makes it possible to notice quality improvement even for rare cases~\cite{barron2019conversation}.
    \item \textit{Informative}. Dataset should contain as many information about each captured image as possible. 
    Precisely the information available during shooting procedure, meta-information about scene properties, information about light sources from different angles, etc.
    \item \textit{Updatable}. Every illumination estimation dataset usually contains ground-truth illumination errors.
    Because of that, the dataset infrastructure should provide simple and reliable way for dataset debugging and tracking of its versions.
    \item \textit{Verifiable}. From the previous point it follows that the dataset should be available for verification, namely all provided markup and ground-truth can be collected and, if necessary, recreated by anyone who just downloads the source images.
    %\todo[inline]{Accuracy Estimation?} I think that this is covered later
    \item \textit{Accessible}. The value of a dataset is decreasing when the downloading process is too complicated or time consuming.
    \item \textit{GDPR compliant}. Even a very good dataset can be of limited use for the European researches if it is not compliant with GDPR, because it may prevent the researchers from publishing some of their results without breaking the regulations.
\end{itemize}

\begin{figure*}[htb]
	\captionsetup[subfloat]{labelformat=empty}
    \centering
    
	\begin{subfloat}[]{}
        \includegraphics[width=0.16\linewidth]{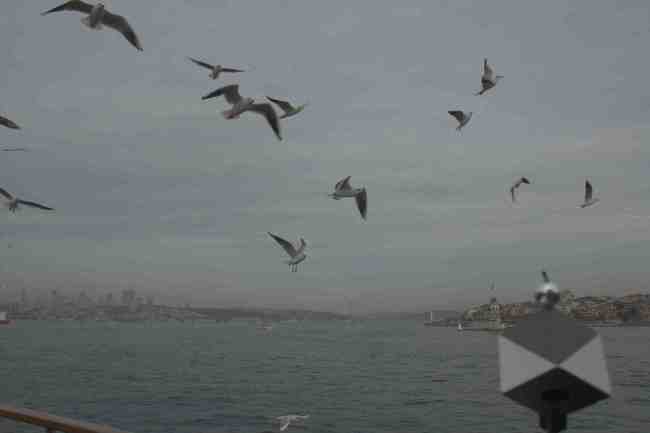}
        \includegraphics[width=0.16\linewidth]{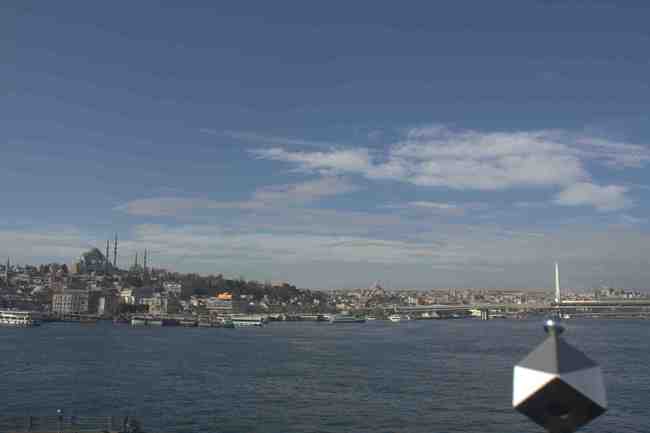}
        \includegraphics[width=0.16\linewidth]{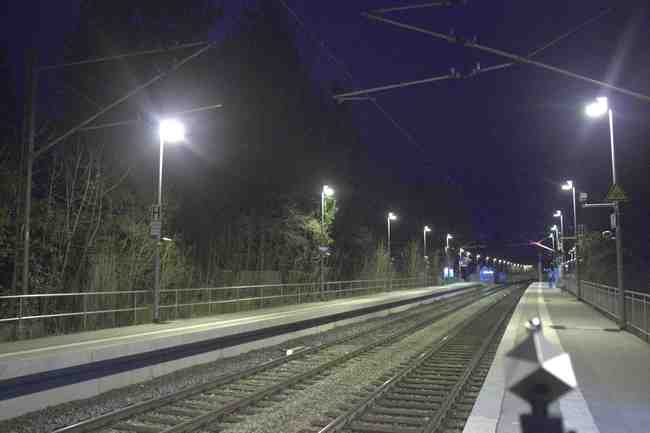}
        \includegraphics[width=0.16\linewidth]{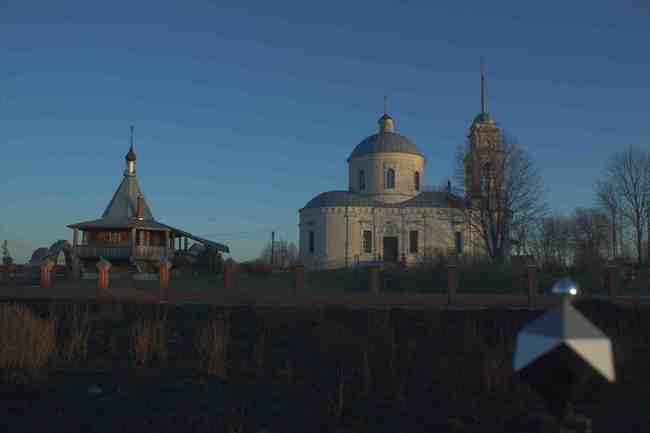}
        \includegraphics[width=0.16\linewidth]{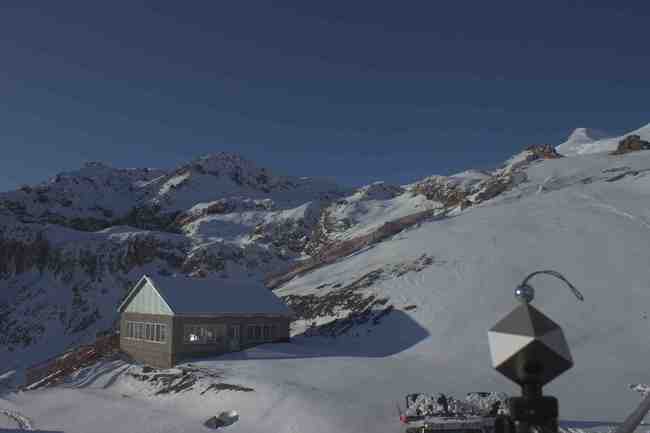}
        \includegraphics[width=0.16\linewidth]{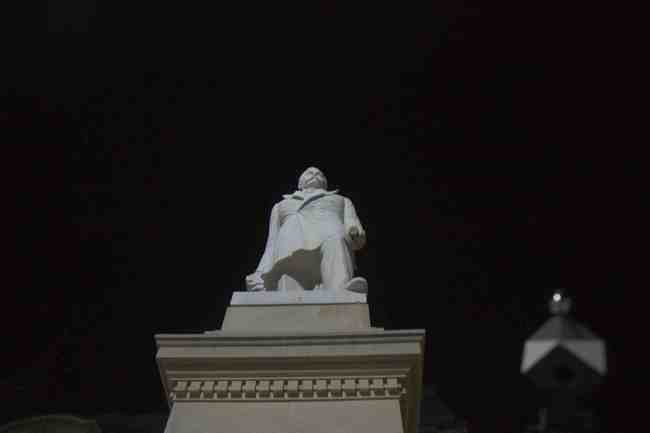}
	\end{subfloat}
	\begin{subfloat}[]{}
	    \includegraphics[width=0.16\linewidth]{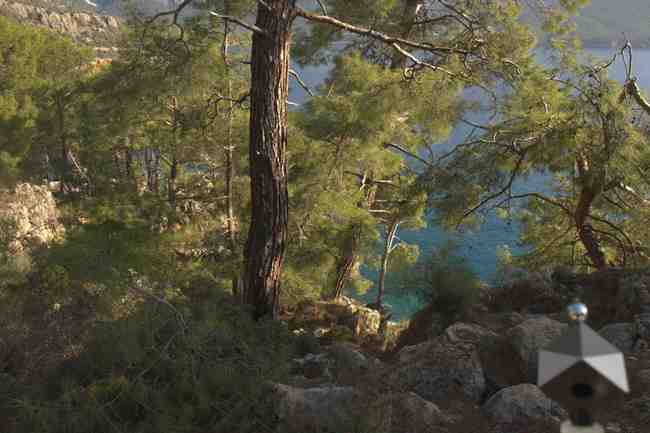}
        \includegraphics[width=0.16\linewidth]{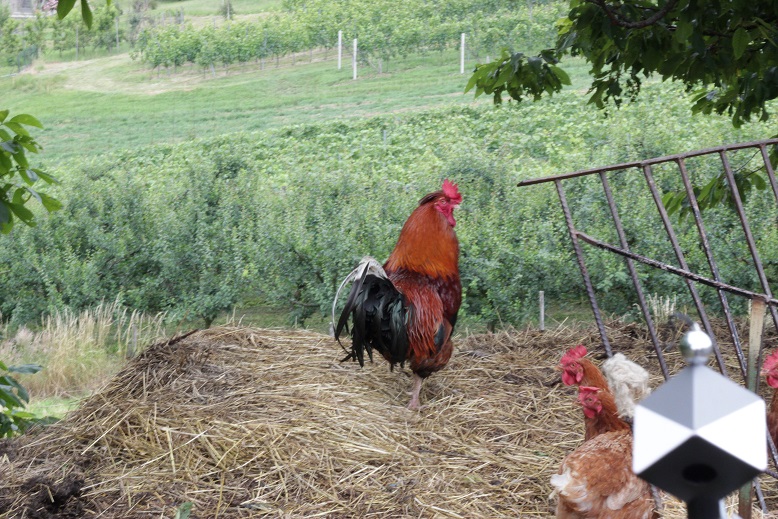} % 00_0368 % cock
        \includegraphics[width=0.16\linewidth]{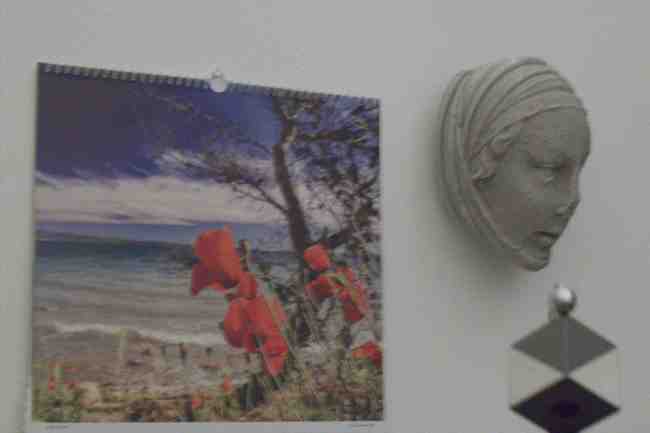}
        \includegraphics[width=0.16\linewidth]{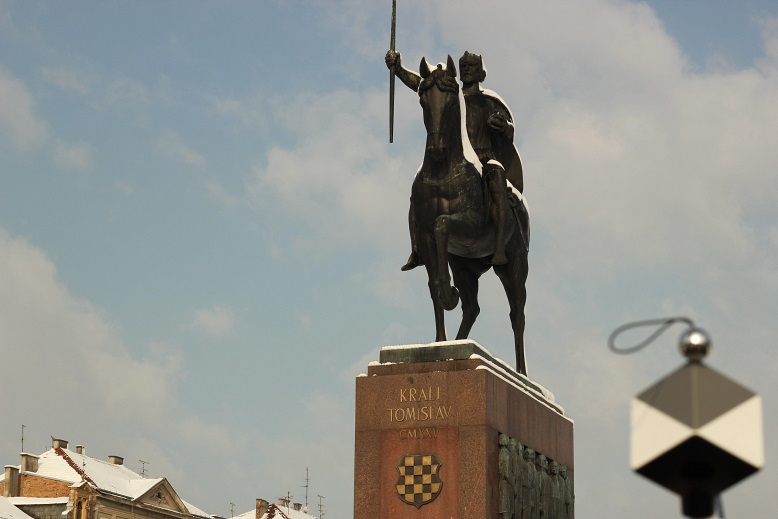} % 00_0501
        \includegraphics[width=0.16\linewidth]{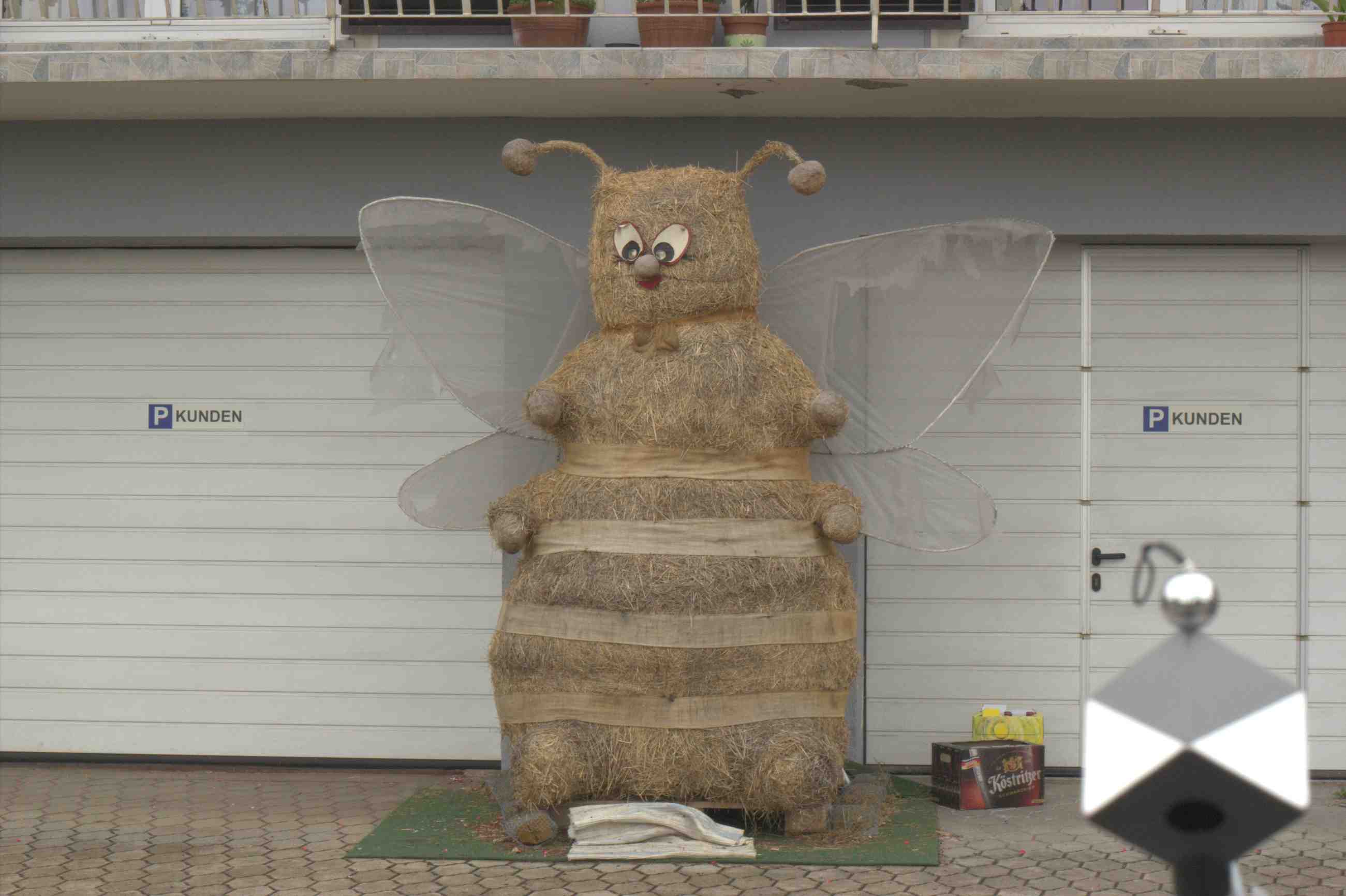} % 00_0177
        \includegraphics[width=0.16\linewidth]{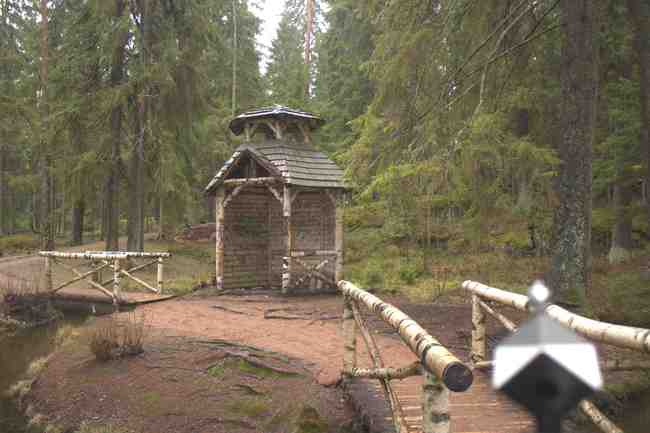}
	\end{subfloat}
	\begin{subfloat}[]{}
        \includegraphics[width=0.16\linewidth]{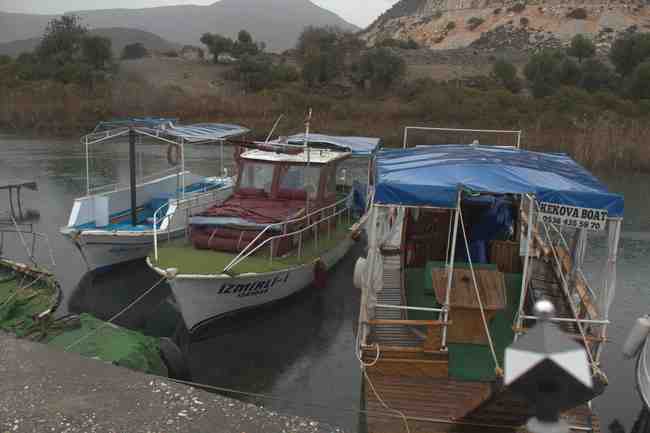}
        \includegraphics[width=0.16\linewidth]{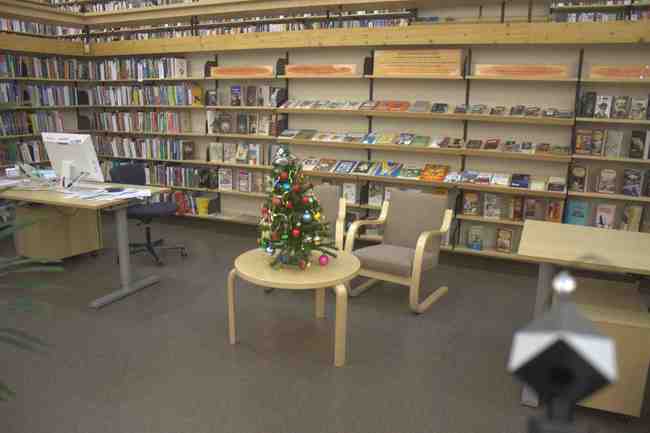}
        \includegraphics[width=0.16\linewidth]{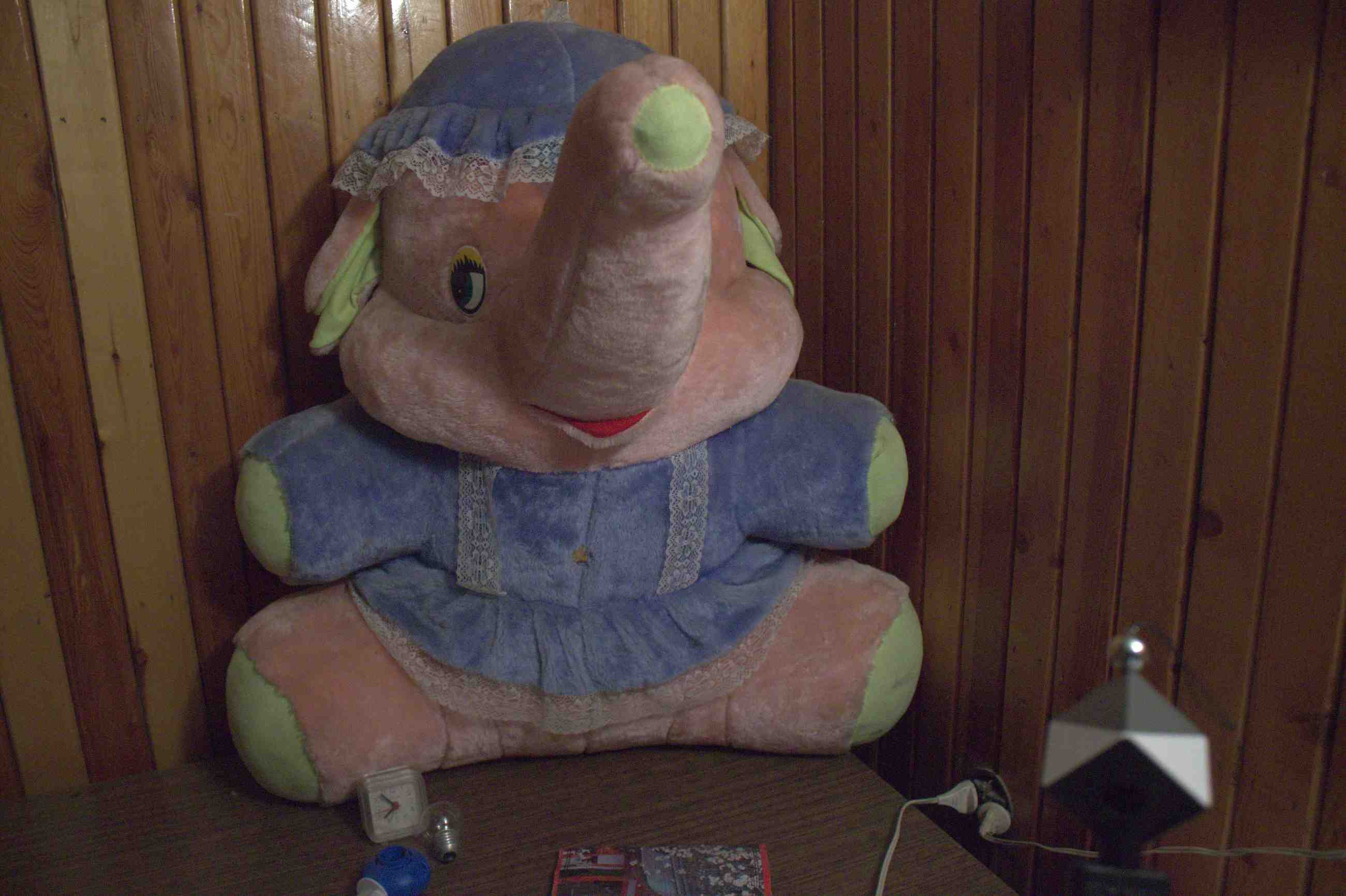} % 05_9565
        \includegraphics[width=0.16\linewidth]{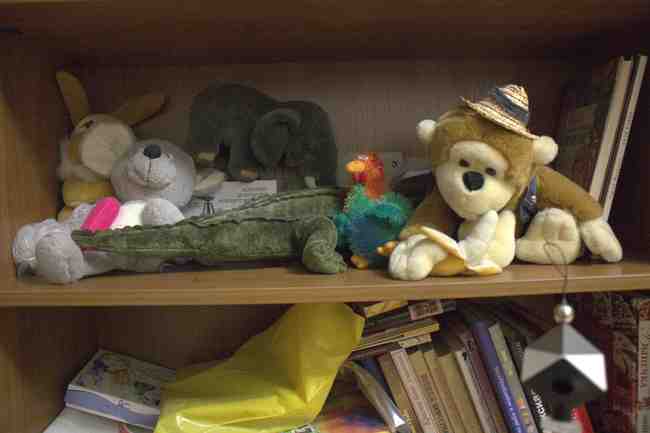}
        \includegraphics[width=0.16\linewidth]{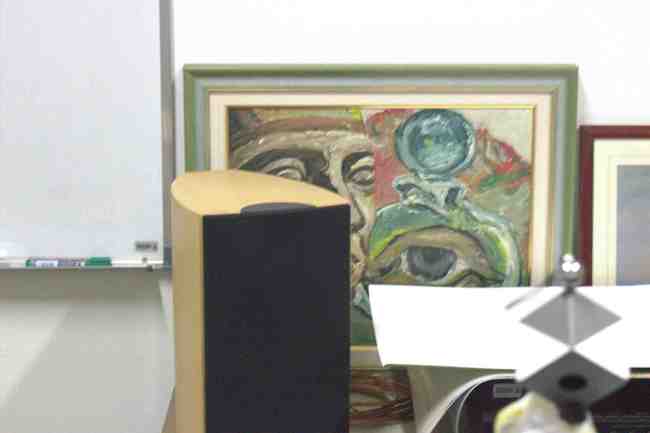}
        \includegraphics[width=0.16\linewidth]{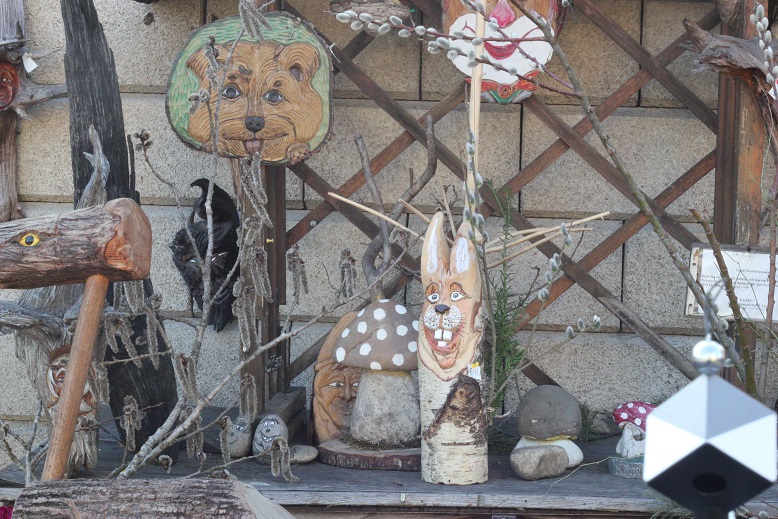} % 00_1638
	\end{subfloat}
	\begin{subfloat}[]{}
		\includegraphics[width=0.16\linewidth]{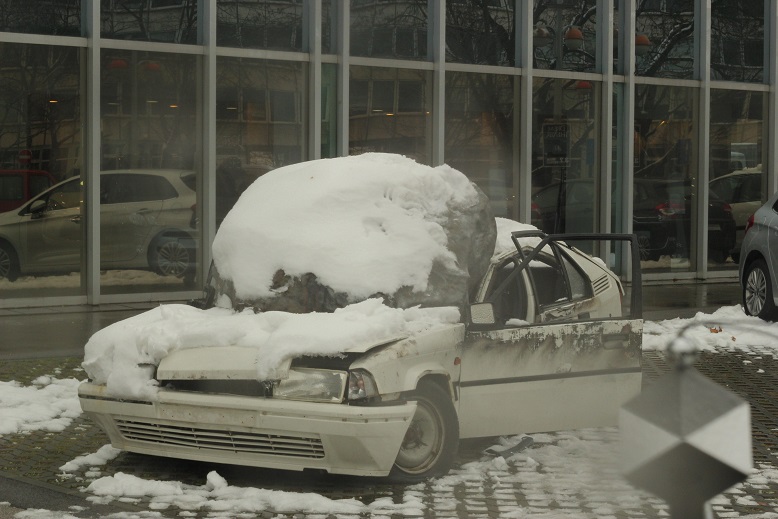} % 00_0021 % car
        \includegraphics[width=0.16\linewidth]{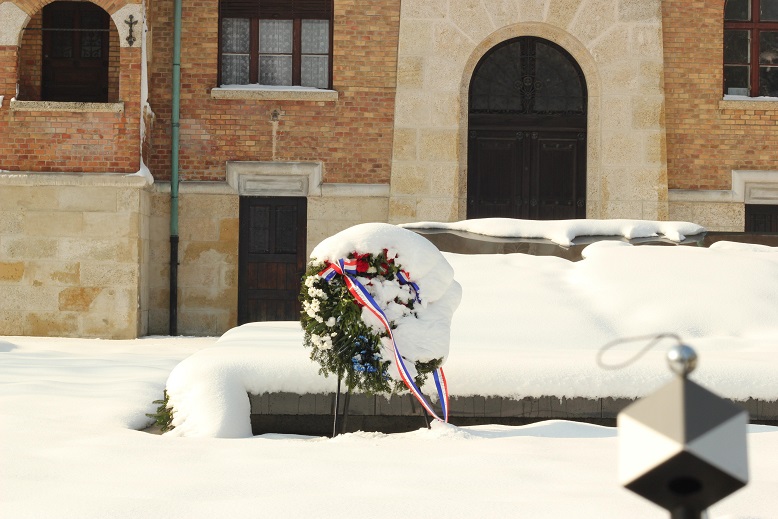} % 00_1628
        \includegraphics[width=0.16\linewidth]{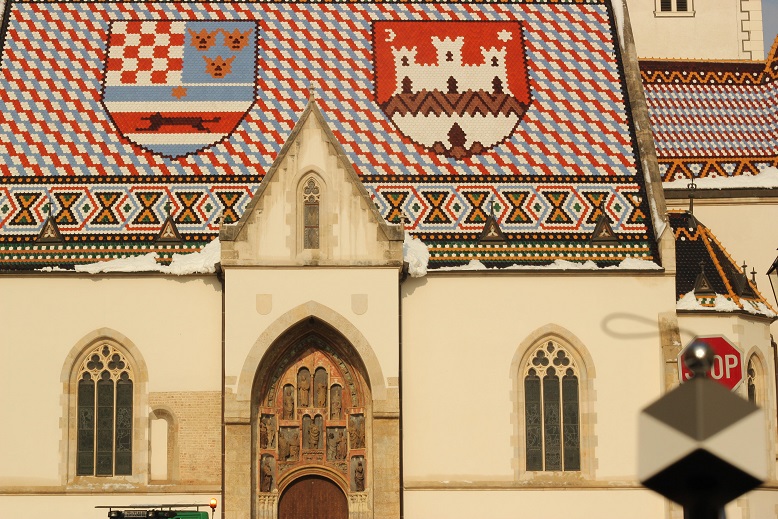} % 00_1001
        \includegraphics[width=0.16\linewidth]{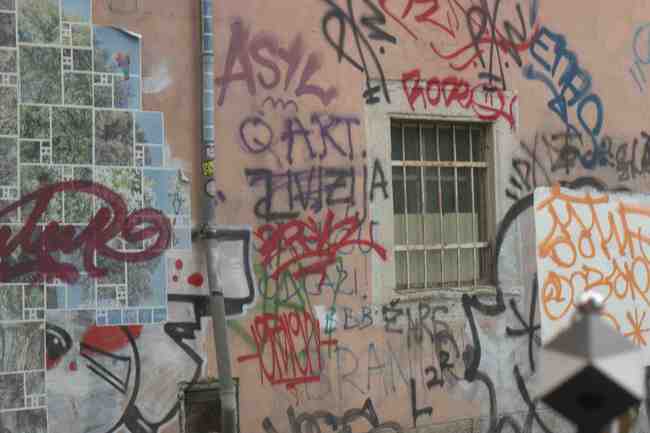}
        \includegraphics[width=0.16\linewidth]{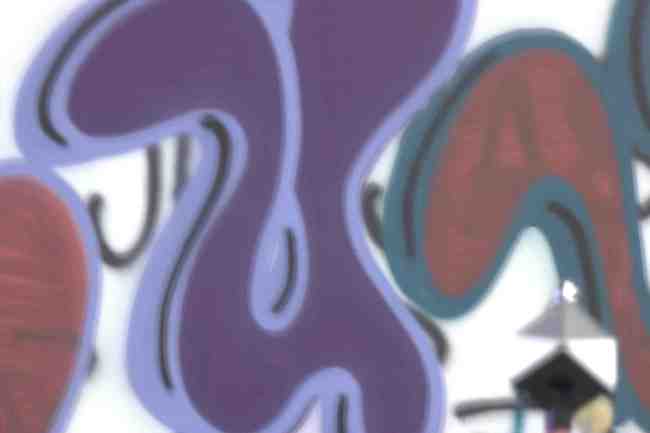}
        \includegraphics[width=0.16\linewidth]{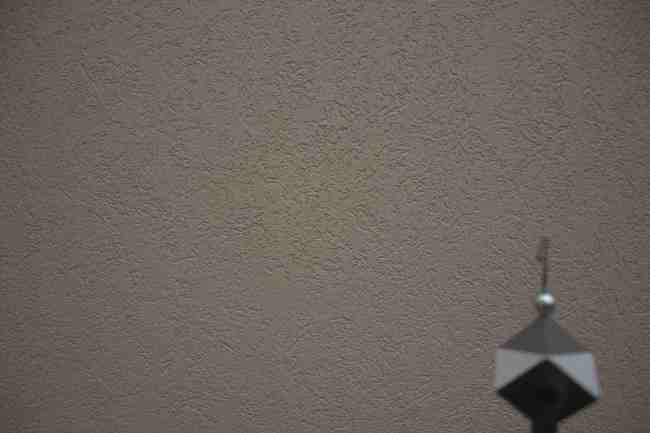}
	\end{subfloat}\\

    \caption{Example images from the newly created Cube++ dataset.}
	\label{fig:cube}
    
\end{figure*}

In this paper a new illumination estimation dataset named Cube++ with all of these properties is described.
It contains 4890 images (see Fig. \ref{fig:cube}) carefully calibrated so as to get highly accurate ground-truth illumination. 
The images were collected in numerous countries, places, and illumination conditions. 
The countries in question include Austria, Croatia, Czechia, Georgia, Germany, Romania, Russia, Slovenia, Turkey, and Ukraine.
In order to enable easy selection of images with specific properties, each image is accompanied by additional semantic information such as whether there are shadows in the image, whether it is an indoor or an outdoor image, whether the scene contains objects with known coloration, etc. 
An example of an image from Cube++ is shown in Fig.~\ref{fig:frontpic}. 
The dataset is appropriate for different light source estimation use cases such as: single light source estimation, two light sources estimation, or estimation of at least one significant light source in the scene.
Finally, some of the collected images were not included in the dataset and they are kept aside to be released later as part of a future illumination estimation benchmark somewhat similar to~\cite{geiger2012we}.

The paper is structured as follows: Section~\ref{sec:existing} describes the most important existing illumination estimation datasets and the problems associated with them, Section~\ref{sec:motivation} gives the motivation for creating a new dataset, Section~\ref{sec:methodology} describes the methodology used to collect the dataset, Section~\ref{sec:proposed} describes the newly proposed Cube++ dataset, Section~\ref{sec:discussion} presents a discussion about the scientific usefulness of contemporary datasets' form and about a potential improvement, and finally, Section~\ref{sec:conclusions} concludes the paper.

% Furthermore, since in most cases the ground-truth illumination for the test partitions of the these datasets is known, problems that may additionally occur include $p$-hacking, using cherry-picked versions of datasets with multiple versions, obtaining the results through hidden hyperparameters not included in the methods' papers, erroneously reported results, exploiting the ground-truth distribution, etc. 
% This can lead to publishing of the results of questionable quality and hence hamper the progress in illumination estimation research. 

%% file: chapters/2_review.tex
\section{Influential existing datasets}
\label{sec:existing}

One of the first illumination estimation datasets with a large number of images was the GreyBall dataset~\cite{ciurea2003large}. 
A gray ball was placed in the scene of each of the $11346$ images to extract the ground-truth illumination. 
% It is interesting to specify these details and put reference on the description of this problem if it is exists
The main problem with this dataset is that the images are non-linearly processed and as such they do not comply with the image formation model given in Eq.~\eqref{eq:image}. 
Furthermore, the images in the GreyBall dataset are relatively small with a size of $240\times 360$.
Finally, the images were extracted from a video that was captured at several locations, which means that many of them have highly correlated illuminations and content. To cope with this problem of high redundancy, it has been proposed to use only a subset of $1135$ images from GreyBall~\cite{bianco2008improving,bianco2010automatic}.

In 2008, the ColorChecker dataset~\cite{gehler2008bayesian} with its $568$ images was published and the ground-truth illumination was extracted by means of putting a color checker instead of a gray ball in the image scenes. 
This dataset was created by two different cameras and its images, which are individually bigger than the ones in the GreyBall dataset, also underwent non-linear processing, which means that similarly as with the GreyBall dataset they are given as 8-bit per-channel JPEG images.

In 2011, the reprocessed version of the ColorChecker dataset that contains only linearly processed images was published~\cite{shi2020online}. 
However, as observed already in 2013~\cite{lynch2013colour}, it was not mentioned clearly enough that the black level was supposed to be subtracted before using the images. 
Despite this observation, a lot of papers continued publishing results of methods obtained on the technically unprepared images with the black level included. 
This effectively led to the circulation of at least three versions of the ColorChecker datasets and the problem was formally addressed in~\cite{finlayson2017curious} by also bringing into question the alleged advances in the illumination estimation research.
In 2018, there was an attempt to rehabilitate the ColorChecker dataset by publishing the recalculated accuracy of various methods by using the allegedly correct ground-truth~\cite{hemrit2018rehabilitating}. 
However, this attempt was marred by serious technical faults and wrong calculations that included comparing the estimations obtained on older versions to the new ground-truth, which only introduced further confusion~\cite{banic2019past}. 
This effectively opens the possibility of more future version of the results on the ColorChecker dataset. 
In short, using the ColorChecker dataset can be very confusing and problematic due to many circulating versions of the alleged results and consequent inappropriate comparisons, and therefore, to avoid problems, it should probably be omitted as the primary dataset choice.

In 2014, nine new NUS datasets with each of them taken by one of nine different cameras were published~\cite{cheng2014illuminant}.
The images were only linearly processed, the black level subtraction was performed from the start in the initial paper, and the number of images was sufficiently high. 
The calibration object used to extract the ground-truth illumination was again a color checker.
However, the problems with the NUS datasets include violations of uniform illumination assumption when having only a single ground-truth illumination, a relatively small number of images per camera with $268$ being the maximum, having the same scenes repeated in images, and not being GDPR compliant as well as neither of the previous datastes.

In 2017, the Cube dataset~\cite{banic2018unsupervised} was published with $1365$ images taken with a single camera and with a SpyderCube~\footnote{\url{https://www.datacolor.com/photography-design/product-overview/spydercube/}} calibration tool used for calibration. 
Due to its geometry that is superior to the one of a color checker, SpyderCube allows for easier detection of the presence of two illuminations and their extraction. 
This was extensively used to carefully calibrate each of the images of the Cube dataset and to obtain an accurate ground-truth. 
Special care was also taken to avoid the violation of the uniform illumination assumption as much as possible. 
The main drawback of the Cube dataset is that it contains only outdoor images, which also negatively affects the ground-truth illumination distribution. 
This drawback was alleviated in the Cube dataset's extension named the Cube+ dataset~\cite{banic2018unsupervised}. 
It contains $342$ additional indoor images for a total of $1707$ images and a wider span of ground-truth illumination distribution similar to the one in other datasets.

A relatively recent dataset is the INTEL-TAU dataset~\cite{laakom2019intel}, a successor to the INTEL-TUT dataset~\cite{aytekin2017data}, with $7022$ images taken by three different cameras. 
While the number of images is sufficiently high, its main drawback is the fact that most of its images do not contain a calibration object in their scenes. 
Namely, it was removed after the initial calibration. 
Although this removes the requirement for masking it out, it also makes it impossible to reliably check and verify the ground-truth calibration and it is known that such errors occur~\cite{zakizadeh2015hybrid}.
Additionally, since the original raw image files are not provided, the EXIF data with the meta-information that may be important to some methods is also not available.
The INTEL-TAU dataset is also completely GDPR compliant. Instead of avoiding problematic scenes, GDPR compliance was achieved by having ''privacy masking applied on all sensitive information'' such as ''recognizable faces, license plates, and other privacy sensitive information''. 
The masking was performed so that ''color component values inside the privacy masking area were averaged''. However, this effectively changes the original content and it may be undesirable in some cases.

A relatively recent dataset is the one for temporal color constancy~\cite{qian2020benchmark}, which contains $600$ sequences of varying length between $3$ and $17$ frames. The dataset has not yet been made publicly available at the moment of writing this paper.

It is also important to mention that in contrast to all the described datasets that contain real-world images taken in mostly uncontrolled conditions, there are a lot of datasets made in fully controlled or even laboratory conditions, such as \cite{barnard2002comparison,geusebroek2005amsterdam,bleier2011color,rizzi2013yaccd2,smagina2020multiple,murmann2019dataset, shepelev2017stereo}.

The main advantage of the laboratory dataset is that it allows to research particular problem in fully-controlled conditions, but the variability of such datasets is often too low.

While other illumination estimation benchmark datasets also exist, it can be argued that the ones mentioned here are the most influential ones. 
They also share many problems with other existing datasets and thus their descriptions also cover most of the problems of other datasets. 
Some characteristics of the datasets mentioned here are summarized in Table~\ref{tab:datasets}.

%% file: chapters/3_motivation.tex
\section{Motivation}
\label{sec:motivation}

After laying out the brief descriptions of some of the best known illumination estimation benchmark datasets, it is possible to identify some of their main problems already recognized by the wider interested research community. 
Therefore, the motivation for creating a new illumination estimation is to try to reduce or entirely eliminate some of the mentioned problems of the existing datasets.

\subsection{Simple technical faults}
\label{subsec:technical}

Probably the most serious and most detrimental problem is the one connected to the technical shortfalls that can happen when creating and publishing a dataset. 
Some of the main such shortfalls are using non-linearly processed images and providing confusing information on black level subtraction.

As for the non-linearly processed images, the solution is to simply avoid performing non-linear processing and this can be simply carried out.

In the case of the black level subtraction, with the earlier datasets this problem occurred due to lack of explicit mentioning of the black level value in the papers that originally described these datasets. 
Additionally, in some cases even a script that demonstrates the proper handling of the black level was either missing or put to a somewhat obscure location. 
In the case of the ColorChecker such problems have led to multiple circulating versions of the ground-truth data and experimental results. 
Therefore, in the case of publishing a new dataset, such and similar problems motivate to clearly provide all necessary details on the required data for the black level subtraction and also to provide an example of how to do it.

\subsection{Reliable ground-truth}
\label{subsec:reliable}

One of the probably least detectable technical faults with serious consequences is erroneous calibration and ground-truth illumination extraction.
Based on the experience with existing datasets, it usually happens that there are multiple illuminations in the scene and the calibration object is under the influence of only one of them, which may not even be the dominant one. 
In that case, even if a method estimates the dominant illumination, it will be penalized because the ground-truth is based on another illumination. 
As mentioned earlier, this was already reported for the ColorChecker dataset.

To make the ground-truth reliable, one should use such a calibration objects that can detect the presence of multiple illuminations.
Examples of the such calibration objects include a gray ball such as in the GreyBall dataset or a SpyderCube instance such as in the Cube+ dataset, because they make it possible to simultaneously observe illuminations coming from different angles, and these can then be checked for any significant difference. An example of capturing two significantly different illuminations with a SpyderCube instance and showing the difference in how they affect color correction is given in~\cite{banic2018unsupervised}. If a significant difference is present, additional steps can be taken to either correctly determine which of the illuminations is the dominant one or to discard the image to prevent any future problems, which finally results in a correctly extracted and reliable ground-truth illumination.

\newcommand\datasetangle{76}

\begin{table*}[h!]
% \normalsize
\small
% \scriptsize
%\tiny
\caption{Characteristics of different illumination estimation datasets; the Cube++ dataset is described later in Section~\ref{sec:proposed}. }
\label{tab:datasets}
\centering
\begin{tabular}{|c|ccccccc|}

	\hline
	&\multicolumn{7}{|c|}{}\\
	%\hline
	\textbf{Characteristic} & \begin{turn}{\datasetangle}GreyBall~\cite{ciurea2003large}\end{turn} & \begin{turn}{\datasetangle}ColorChecker~\cite{shi2020online}\end{turn} & \begin{turn}{\datasetangle}NUS~\cite{cheng2014illuminant}\end{turn}  & \begin{turn}{\datasetangle}Cube+~\cite{banic2018unsupervised}\end{turn} & \begin{turn}{\datasetangle}INTEL-TAU~\cite{laakom2019intel}\end{turn} & \begin{turn}{\datasetangle}TCC~\cite{qian2020benchmark}\end{turn} & \begin{turn}{\datasetangle}\textbf{Cube++}\end{turn} \\ % put a reference
	\hline
	Number of images              & 11346 \footnotemark\global\let\saved  & 568                                 &  1736       & 1707       & 7022         & 600 sequences                        & 4890       \\
	Number of sensor types        & 1                                     & 2                                   &  8          & 1          & 3            & 1                                    & 1          \\
	GDPR compliance               & -                                     & -                                   &  -          & -          & \checkmark   & \checkmark                           & \checkmark \\
	Undistored content            & \checkmark                            & \checkmark                          &  \checkmark & \checkmark & -            & \checkmark                           & \checkmark \\
	Color target in the scene     & \checkmark                            & \checkmark                          &  \checkmark & \checkmark & -            & \checkmark                           & \checkmark \\
	Color target type             & gray ball                             & color checker (CC)                  &  CC         & SpyderCube & CC           & SpyderCube                           & SpyderCube \\
	Meta-info about the scene     & -                                     & -                                   &  -          & -          & -            & -                                    & \checkmark \\
	Night images                  & -                                     & -                                   &  -          & -          & -            & -                                    & \checkmark \\
	Winter season                 & -                                     & -                                   &  -          & -          & -            & \checkmark                           & \checkmark \\
	Year                          & 2003                                  & 2020 \footnotemark\global\let\saved &  2014       & 2019       & 2020         & 2020  \footnotemark\global\let\saved & 2020       \\
    License                       & -                                     & -                                   &  -          & -          & CC BY-SA 4.0 & MIT                                & CC BY 4.0 \\

    % License                       & -                                     & -                                   &  -          & -          &  \includegraphics[scale=0.5]{img/licenses/cc-by-sa.png} & N/A                                  & \includegraphics[scale=0.5]{img/licenses/cc-by.png}  \\

    \hline

\end{tabular}
\end{table*}

\subsection{Verifiable ground-truth}
\label{subsec:verifiable}

While the ground-truth should primarily be reliable, it should also be verifiable in order to add an additional layer of reliability. The simplest way of making the ground-truth of a dataset verifiable is to have all the dataset images contain a calibration objects in their scenes. In that way the ground-truth can easily be extracted by other researchers and then compared to the originally provided one to look for potential errors. Additionally, the very visual information can help identify cases such as e.g. having the calibration object in a shadow while the majority of the scene is outside of that shadow.

\subsection{Content variety}
\label{subsec:content}

A new illumination estimation dataset should have a high content variety. 
While this seems rather obvious, it is not always put into practice to the full extent. 
For example, while the GrayBall dataset contains over $11$k images, they are highly correlated and thus effectively not as rich in content as it may seem at first. 
In the case of datasets such as the ColorChecker dataset or the NUS datasets, all images were taken at the same geographical location and during the same season. 
None of the images there were taken e.g. during winter or at night. 
Such content choice restriction results in failure to cover many interesting and challenging environments that illumination estimation methods encounter in real-world applications and that should also be included in the research.

\subsection{Illumination variety}
\label{subsec:illumination}

Having an appropriate ground-truth illumination variety in an illumination estimation dataset is important for several reasons.
The most important one is to closely cover as much as possible of the illuminations that are encountered in the real-world applications because in that way the illumination estimation methods can be properly trained and tested. 
In most existing datasets the chromaticities of the illuminations are rarely too far away from the Planckian locus~\cite{tregenza2013design}. 
This means that less common colors of the artificial illumination sources are also effectively excluded. 
Therefore, another thing to consider when creating a new illumination estimation dataset is the inclusion of such less common ground-truth illuminations.

An additional reason to have a sufficient ground-truth illumination variety is to prevent abuses of some often used error statistics that are possible if the ground-truth illumination are too clustered~\cite{banic2015perceptual}. 
Such abuses can lead to false conclusions about the performance of the tested methods and consequently be detrimental for the research community and practitioners.

\subsection{Checking for multiple illuminations}
\label{subsec:mutliple}

The majority of the illumination estimation datasets provide only a single ground-truth illumination per image. This effectively means that in terms of evaluation these datasets implicitly assume an uniform illumination. However, it is know that in illumination estimation datasets this is generally not the case~\cite{banic2019past}. As a matter of fact, any image with shadows has already effectively at least two illuminations that may differ significantly and this can also have a significant outcome on the later color correction step~\cite{banic2018unsupervised}. Additionally, even if there are no shadows, it is still possible for an image to be under the influence of multiple illuminations. In that case having a calibration object that is designed to successfully capture the illumination from only one direction at a time will fail to capture all the illuminations in the scene, let alone to detect their presence. Capturing only a single illumination when there are more present also leads to a problem during the evaluation. Namely, if a method correctly estimates one of the illuminations, but the other one is marked as the ground-truth, it may be argued that in this case the method is being unfairly judged. Because of that, an illumination estimation dataset should preferably use calibration objects that can simultaneously capture the illumination color from multiple directions. This would solve at least two problems. First, it would detect whether there are multiple illuminations in the first place, and second, if there really are multiple illuminations in the scene, then such a calibration object will capture more information on them. An example of such a calibration object is the SpyderCube object that has been described earlier.

\subsection{Number of images}
\label{subsec:number}

While some of the previous datasets with non-linearly processed images are obviously disadvantageous, some of them like the GreyBall dataset have the advantage of having thousands of images, which still makes them attractive to many researchers. 
Therefore, besides having a technically correct dataset, it is also important to make it have a sufficiently large number of images. 
This can result in both making the dataset desirable by offering a lot of useful data as well as simultaneously discouraging researches from using the inferior older datasets just because of their size. 
As for how large exactly a dataset should be, it should contain several thousands of images to outsize the existing datasets of lower quality and also to enable new breakthroughs.
Finally, having a dataset with a large number of images is a prerequisite for achieving the previously mentioned content and illumination variety.

\addtocounter{footnote}{-2}
\footnotetext{GreyBall and TCC images are taken from the video, so they are highly correlated. }
\addtocounter{footnote}{1}
\footnotetext{The cited version of ground truth values were published in 2020, the original dataset was published in 2008.}    
\addtocounter{footnote}{1}
\footnotetext{Expected in 2020, the data is not published at the time of writing. } 

\subsection{Semantic data}
\label{subsec:semantic}

In numerous cases, additional semantic information can be useful for research of specific illumination estimation methods.  For example, some of the methods may be interested in being explicitly trained only on indoor or outdoor images.  Others may be interested in training images that contain no shadows whatsoever since they introduce additional illuminations. More generally, it may be useful to know whether there is a violation of the uniform illumination assumption on a given image. In such cases, it can be highly practical to be able to efficiently filter out images from a dataset based on some given criteria.

Because of that, a useful addition to a new illumination estimation benchmark dataset would be semantic information for each image. In that way, the research could be speeded up by not requiring researchers to label the images from scratch. Additionally, if such semantic information were given in advance, a lot of potential label mismatches between the labels created by different researchers could also be prevented.

\subsection{Privacy concerns}
\label{subsec:privacy}

With the recent arrival of regulations such as the General Data Protection Regulation~(GDPR), it becomes ever more important to respect privacy in publicly available images. 
This also means that using images from previous datasets with e.g. recognizable people or registration plates may nowadays be potentially seen as problematic. 
With respect to this, for the sake of respecting privacy, any new illumination estimation dataset should also take care of avoiding images that would contain any content that could compromise someones privacy.

On the other hand, if a public dataset is also supposed to be useful for development of methods that rely on e.g. faces~\cite{bianco2012face,knorr2014real} or sclera~\cite{males2017colour}, then it should obviously also contain images with faces. 
However, in that case it would be appropriate to obtain the consent for public use from the persons present in the image scenes. 
That would enable the researchers to use and show these images publicly in papers.

\subsection{Multiple instances of the same sensor class}
\label{subsec:sensors}

There can be significant differences between spectral characteristics of different sensors used by various cameras. This effectively means that a learning-based method that has been successfully trained on the images created by a camera of one model will not necessarily perform well on images created by a camera of another model without some adjustments. As a result, the problem of inter-camera color constancy has recently started to gain ever more     attention~\cite{banic2018unsupervised,bianco2019quasi,afifi2019sensor}. Since almost every dataset was taken with another camera sensor, there is no shortage of training and testing images.

On the other hand, it is known and it can be shown that even for the instances of the same sensor class there are measurable differences in the spectral characteristics~\cite{koskinen2020cross}. Hence, to check the significance of the impact of these differences on the accuracy of illumination estimation methods, an interesting feature of an illumination estimation dataset would be to have images created by several instances of the same sensor class. In addition to ground-truth illumination, such a dataset would also have to provide the sensor instance labels for each image.

%% file: chapters/4_methodology.tex
\section{Acquisition methodology}
\label{sec:methodology}

By identifying the problems with the existing datasets and describing some desired properties of the future datasets, the guidelines for creating a new illumination estimation dataset have been laid out. One of the main goals of this paper is not just to provide a theoretical framework, but also to create and propose a dataset by following these guidelines. The first step in doing so is to describe the used acquisition methodology.

\subsection{Technical setup}
\label{subsec:setup}

\subsubsection{Color target (SpyderCube) characterisation}

\begin{figure}
    \centering
    \includegraphics[width=\linewidth]{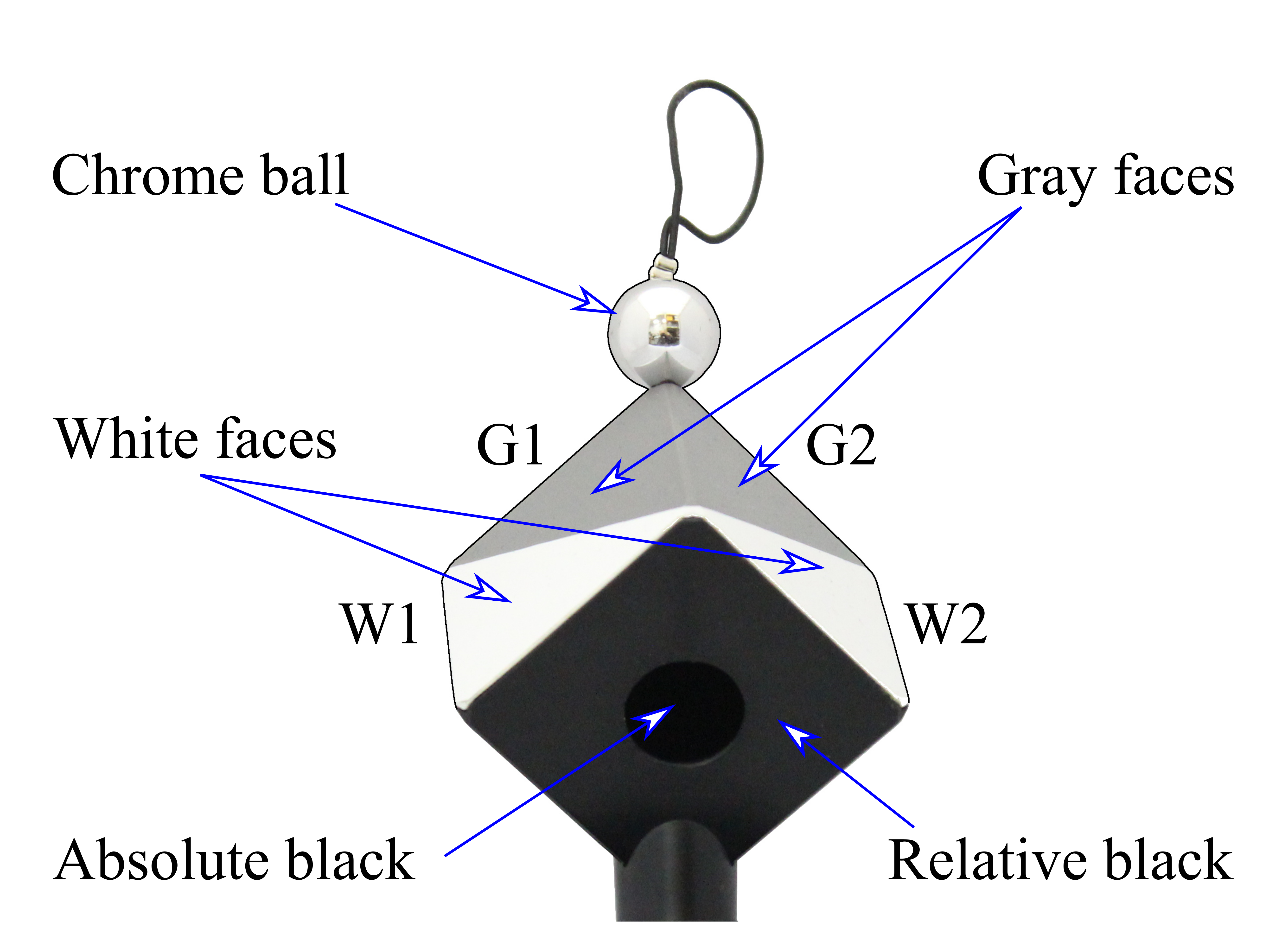}
    \caption{SpyderCube general look.} 
    \label{setup:sc_general_view}
\end{figure}

As the calibration tool in the newly proposed dataset, the SpyderCube instances were used. SpyderCube is a color target for photographers whose main purpose is to help them to adjust the white balance manually.
The general look of the SpyderCube is given in Fig.~\ref{setup:sc_general_view}. A chrome ball is used to analyse specular highlights, the white on two faces is used to estimate true highlight value, the gray on two faces represents the midtone of the image and its color temperature, and the bottom black face is used to evaluate shadow values in the scene in relation to the black trap i.e. the hole, which represents absolute black.

According to the manufacturer company Datacolor, the gray cube faces are neutral gray with a reflection coefficient of 18\%.
%The coefficient of reflection of the white halves of the cube faces is not given.
%I would not use the sentence about the white, it should be 100% by definition: https://pixelsandwanderlust.com/what-is-middle-grey-understanding-18-grey-reflectance/

Since SpyderCube is a relatively low-cost tool, some doubts about its declared optical properties could arise. To validate its properties, two SpyderCube instances, labeled SC1 and SC2, were compared. Individual faces of these SpyderCube instances were named G1, G2 and W1, W2, as shown in Fig.~\ref{setup:sc_general_view}.

\begin{figure}
    \centering
    \includegraphics[width=\linewidth]{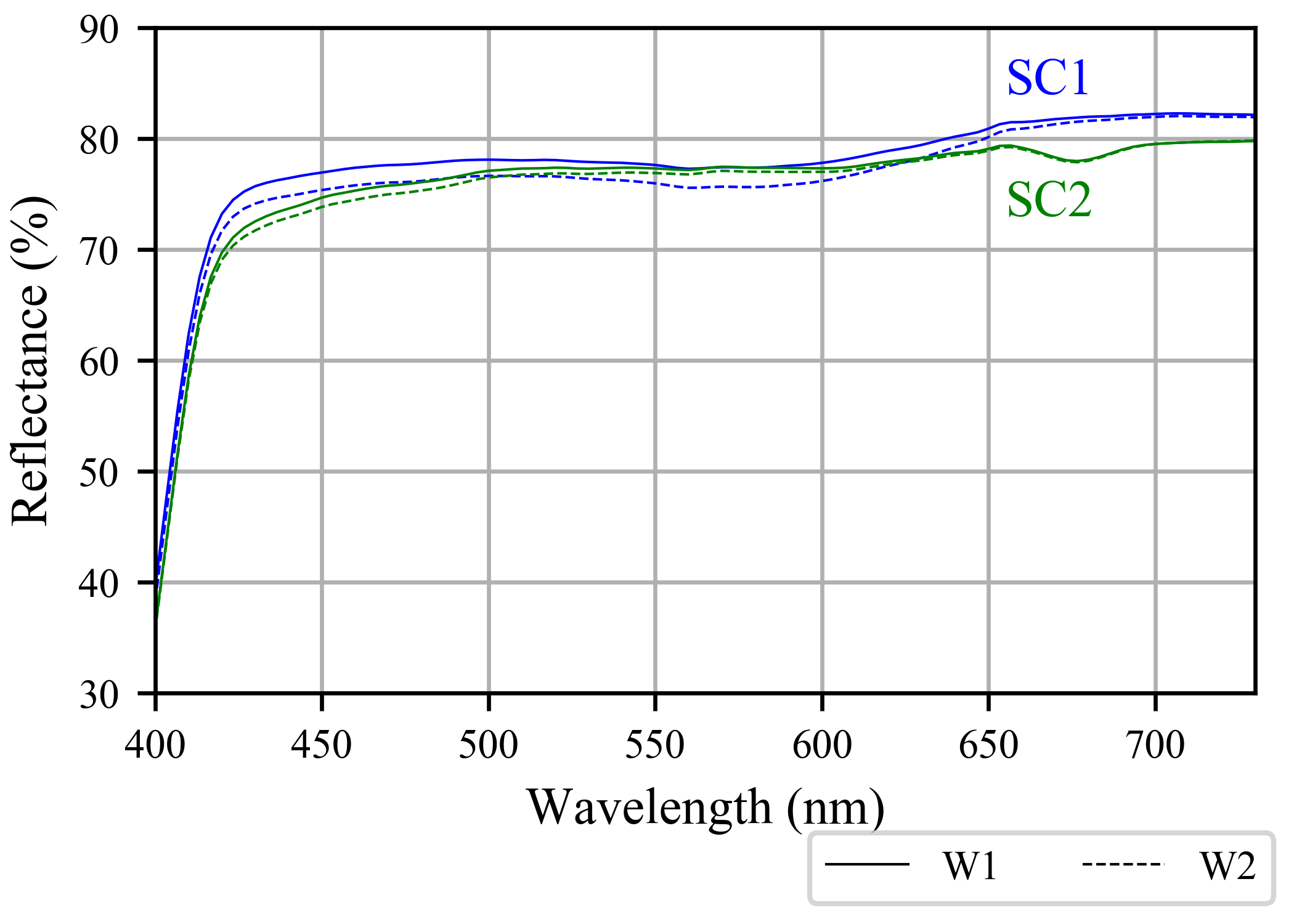}
    \caption{SC1 and SC2 white parts reflectance spectra.}
    \label{setup:sc_white}
\end{figure}

\begin{figure}
    \centering
    \includegraphics[width=\linewidth]{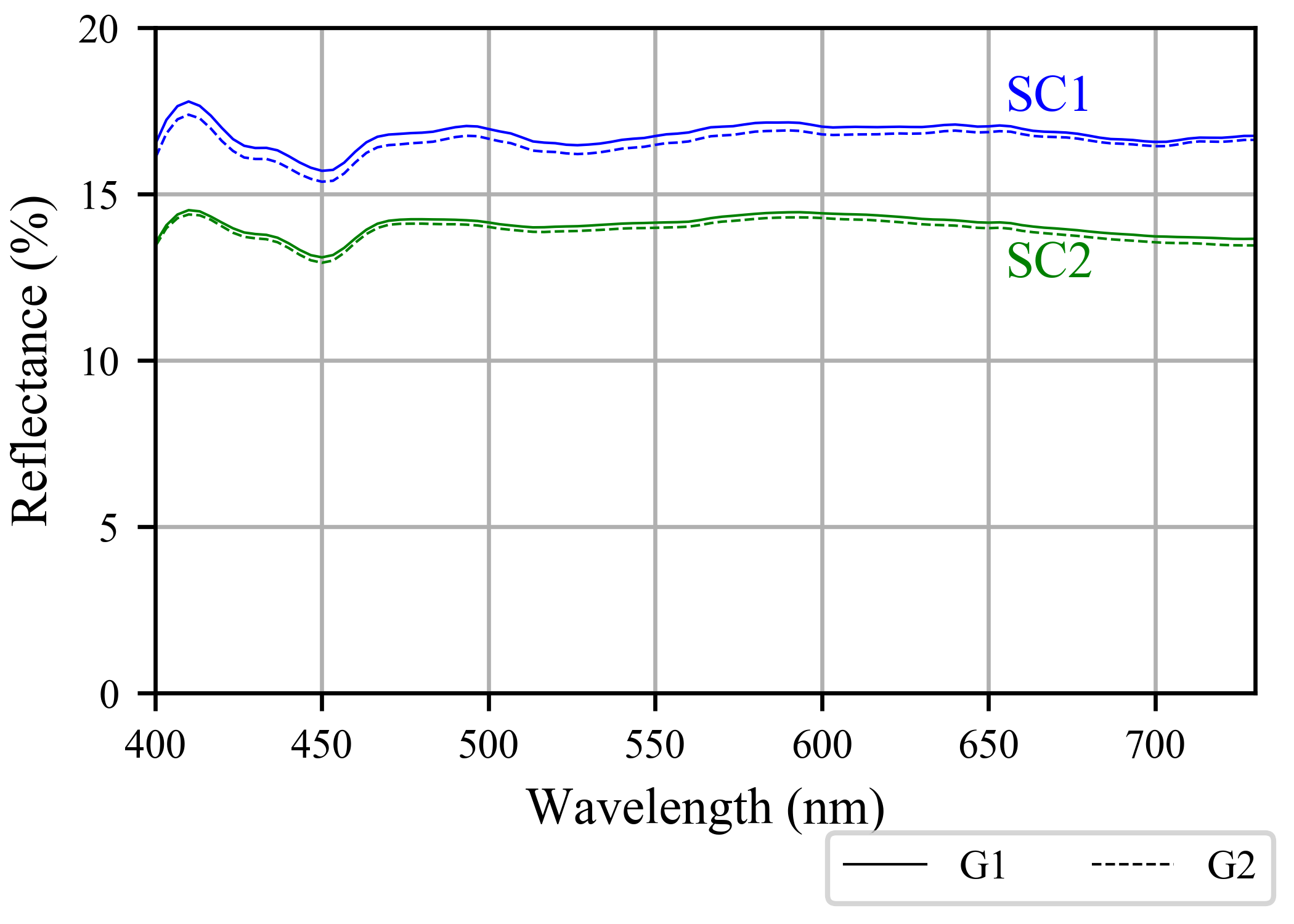}
    \caption{SC1 and SC2 gray parts reflectance spectra.}
    \label{setup:sc_gray}
\end{figure}

Reflection spectra of SpyderCube parts were measured using a Eye-One Pro spectrophotometer by X-Rite in high-resolution mode of 3.3 nm with the help of the spotread utility from Argyll CMS\footnote{\url{http://www.argyllcms.com/}}. For each part, three measurements were made and the results were averaged. Figures~\ref{setup:sc_white} and~\ref{setup:sc_gray} show the spectral reflection coefficients of the white and gray parts of the SC1 and SC2, respectively.

These measurements lead to the following observations:%The following circumstances are noteworthy in the above diagrams:
\begin{itemize}
    \item Gray parts of both SpyderCube instances are not ``ideal'' gray, i.e. the reflection spectra slightly depend on the wavelength.
    The sensitivity of the blue sensor in many cameras has a maximum at around 450 nm wavelength, and the reflection coefficients of gray parts G1 and G2 have a noticeable drop in the blue band.
    \item Each SpyderCube instance has small differences between reflection coefficients of its own gray parts G1 and G2.
    \item There are rather big differences between the gray parts reflection coefficients of the two measured SpyderCube instances.
    \item White parts of both SpyderCube instances are also not ``ideal'' white, i.e. the reflection spectra are not horizontal lines.
    \item Differences of the white parts W1 and W2 reflection coefficients of the both SpyderCube instances are small.
\end{itemize}

The idea behind SpyderCube as a calibration tool is that it does not distort the color of the illumination source, i.e. it is assumed to be ``color neutral''. 
From this point of view, what is important is the similarity between the shapes of the curves of reflection coefficients for the two SpyderCube instances, and not the differences between the curves' values.
From the measurements it can be concluded that the curve shapes are indeed very similar. 
Therefore, the SpyderCube ``color neutrality'' assumption generally holds with one exception being the blue region of the spectrum for the white faces.

The degree of SpyderCube's color neutrality is one of the most important factors for accurate ground-truth extraction.
The height of the grey reflection coefficient curve does not significantly influence the ground-truth extraction.
Compared to other uncertainties, the measured deviations from color neutrality have only rather a small impact on the ground-truth.

Nevertheless, to measure the amount of this impact in terms of practical use, several images of two SpyderCube instances that were simultaneously in the same scene were captured with a Canon 600D camera under a D50-like illumination. 
The average difference between the ground-truth illuminations extracted from the faces of each SpyderCube instance and measured in terms of the angular reproduction error~\cite{finlayson2016reproduction} was about $0.15^{\circ}$, which is in terms of color reproduction insignificant and invisible~\cite{finlayson2005colour}. 

Performed measurements effectively demonstrate that using grey faces of different SpyderCube instances in different images has no significant effect on the overall ground-truth extraction quality.
Still, SpyderCube quality should be studied in details also for all types of complicated artificial light sources (such as gas discharge lamp, etc.).

\subsubsection{Handheld setup}

\begin{figure}
    \centering
    \includegraphics[width=\linewidth]{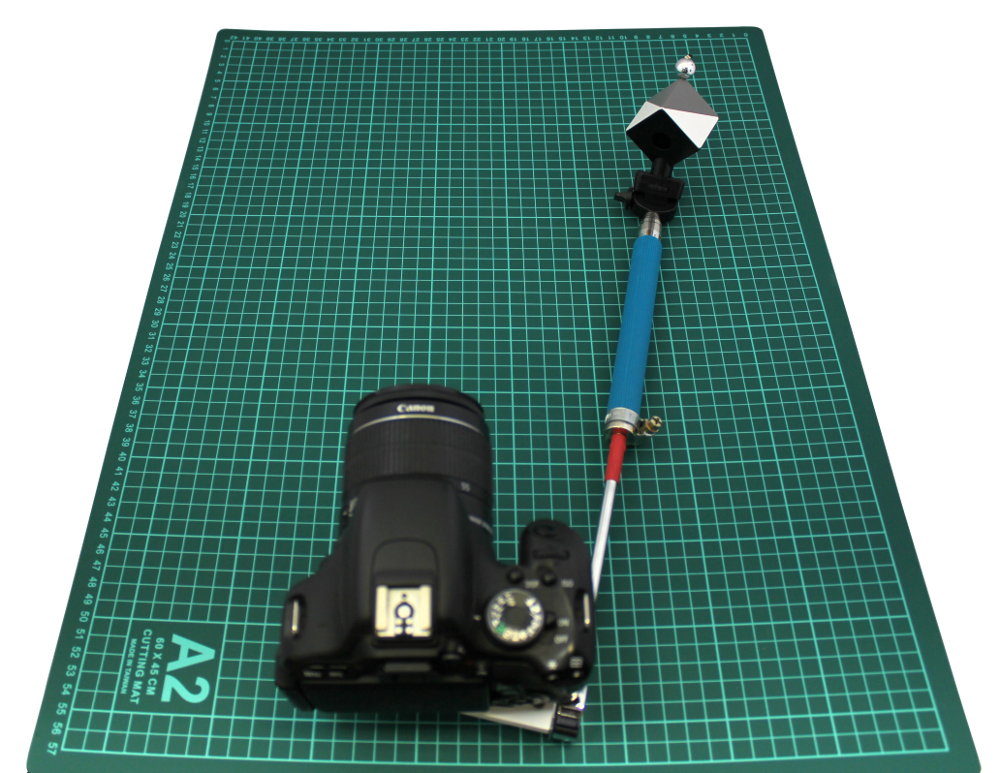}
    \caption{The general look of the handheld setup with Canon 600D camera.}
    \label{setup:handheld_setup}
\end{figure}

To collect the dataset in natural conditions, the following equipment shown in  Fig.~\ref{setup:handheld_setup} was used:

\begin{itemize}
    \item Canon 550D camera or Canon 600D,
    \item SpyderCube calibration tool, and
    \item special attachment of the cube to the camera. 
\end{itemize}

%Two sets of equipment were used.
Special cube fasteners were built that allow the cube to be positioned so that it appears near the lower right corner of the image. The fasteners can also be rotated both in horizontal and vertical planes. 
The distance of the cube from the camera can be adjusted using a telescopic mono-pod and during the dataset images capturing it was set to 50~cm.
The experience gained while collecting the dataset images has led to the conclusion that the custom-built handheld setup is convenient to use.

% =========================================================

\subsection{Data collection and filtration}
\label{subsec:collection_process}

\begin{figure}
    \begin{center}
    \subfloat[]{
    \includegraphics[width=0.48\linewidth]{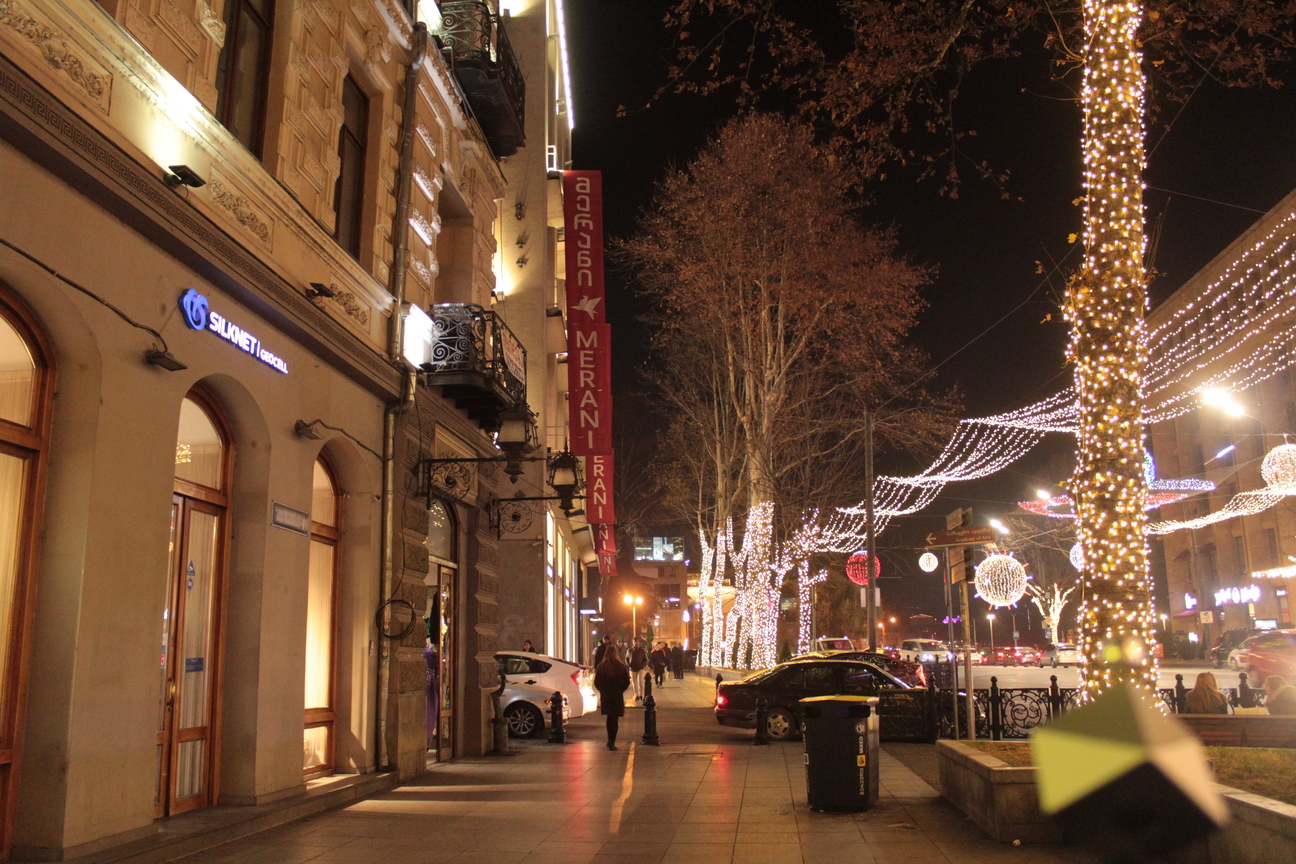}
    \label{fig:not_same_illumination_1}
    }
    % \hfill
    \subfloat[]{
    \includegraphics[width=0.48\linewidth]{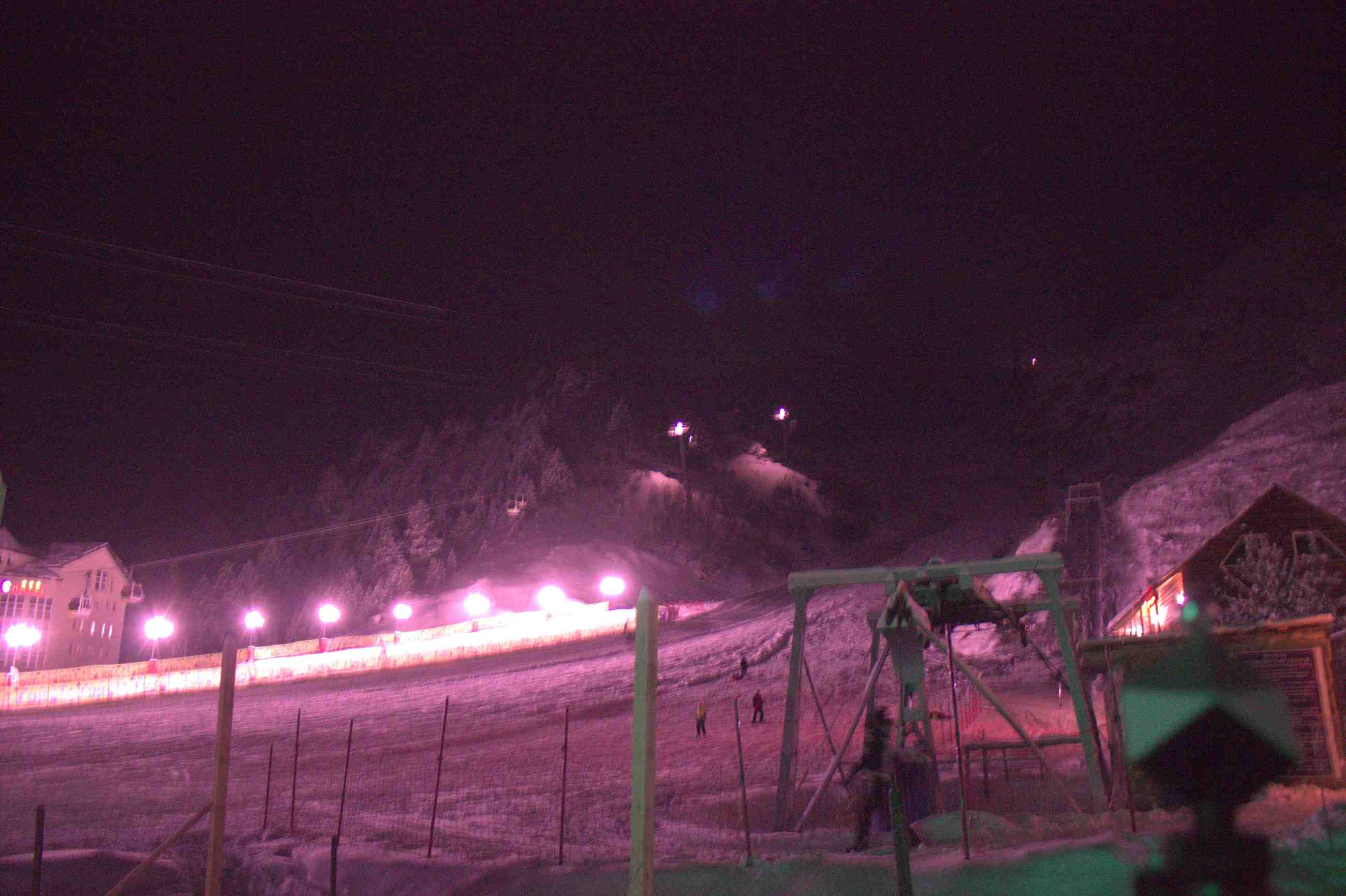}
    \label{fig:not_same_illumination_2}}
    \hfill
    \subfloat[]{
    \includegraphics[width=0.48\linewidth]{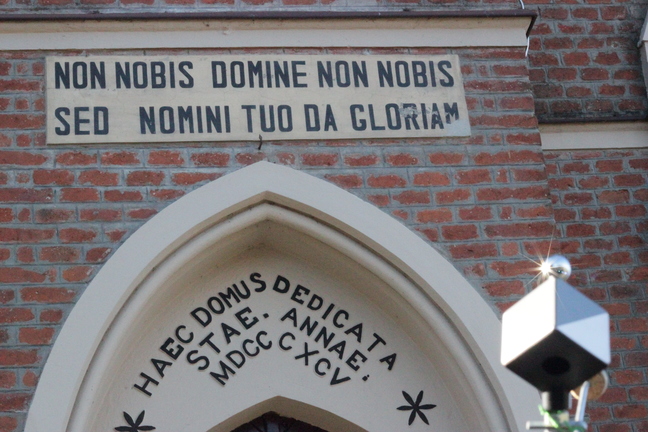}
    \label{fig:overexposed}
    }
    % \hfill
    \subfloat[]{
    \includegraphics[width=0.48\linewidth]{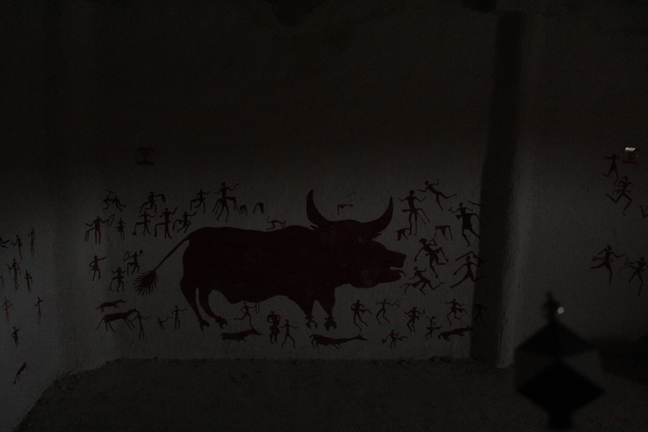}
    \label{fig:dark}
    }
    \end{center}
    \caption{Examples of images that should be excluded from the dataset: a) the color target is illuminated by the local lantern from the near shop, the color is different from the lighting of the most of scene; b) the color target is illuminated by sources that have almost no effect on the lighting of the observed scene; c) overexposed color target; d) the overly dark image.} 
    % c) presence of the glare on the one side of the cube target;
     \label{data_collection:examples}
\end{figure}

The main thing to pay attention to during the image capturing was to assure that the used target cube and the majority of the observed scene are under the same illumination or illuminations.
Examples of images with scenes where this requirement was not met are shown in Fig.~\ref{data_collection:examples}.%#.~\ref{fig:not_same_illumination_1} and~~\ref{fig:not_same_illumination_1}.

Another significant factor that prevents accurate ground-truth extraction is the occurrence of  glare on the color target. 
Images with this issue are usually characterized by clipping of the values in one of the color channels on the gray or white faces of the color target. 
The overexposure can be avoided in at least two ways: either by using manual camera settings or by specifying relative exposure compensation. % for camera automatic estimation. 
Dimming by one step usually turned out to be enough during the image collecting. 
Manual camera settings and one step lighting can also help to properly deal with the overly dark images.
Examples of an overexposed and a too dark image are shown in Fig.~\ref{fig:overexposed} and Fig.~\ref{fig:dark}, respectively.

It should again be mentioned that there may often be several different illumination sources in one scene, commonly two, e.g., sun and sky or sky and streetlight. 
In this case, especially if the areas of the scene parts illuminated by different light sources are comparable, it is preferable to place the cube so that both illumination colors are captured by different cube faces.
By doing so, it is later possible to simultaneously extract the illumination color of both influential scene light sources. 

One of the main problems during image acquisition was to find the right position for the photographer to avoid differences between illuminations influencing the target cube and most of the observed scene. 
A lot of interesting scenes are available only in urban areas where there are a lot of different artificial illuminations. 
However, scenes in urban areas are usually full of different personal data like faces or plate numbers, which means that there are some difficulties related to GDPR. 
To make Cube++ GDPR compliant, the images with humans in the scene were filtered out and removed. 
This was done both automatically by using YOLOv3~\cite{redmon2018yolov3} and manually by additionally checking each image. 

% During the final quality filtration, all images were divided into three categories: a)~\textbf{good images} or images with full light source estimation, where the cube was illuminated by all the main light sources in the scene, tag in markup: \textbf{full\_estimation}:true, b)~\textbf{bad images} where the cube does not allow to determine the illumination rating consistent with the scene, and c)~\textbf{the rest i.e. difficult}, tag in markup: \textbf{full\_estimation}:false.
% As a result, about 400 images were marked as bad and removed from the dataset, about 1000 images were marked as difficult, and the rest of the images were marked as good.

\begin{figure}
    \begin{center}
    \subfloat[]{
    \includegraphics[width=0.48\linewidth]{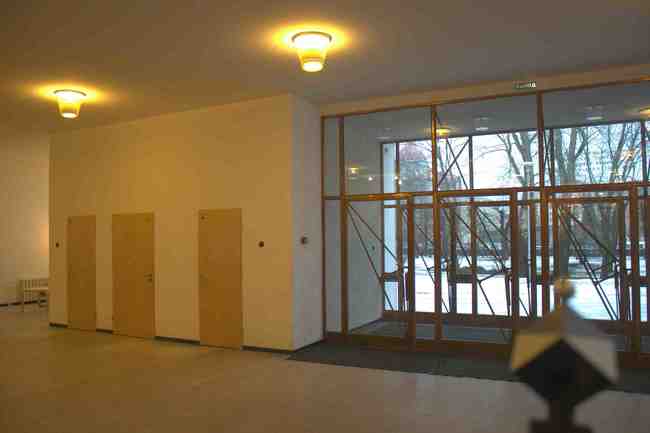}
    \label{fig:has_other_1}
    }
    % \hfill
    \subfloat[]{
    \includegraphics[width=0.48\linewidth]{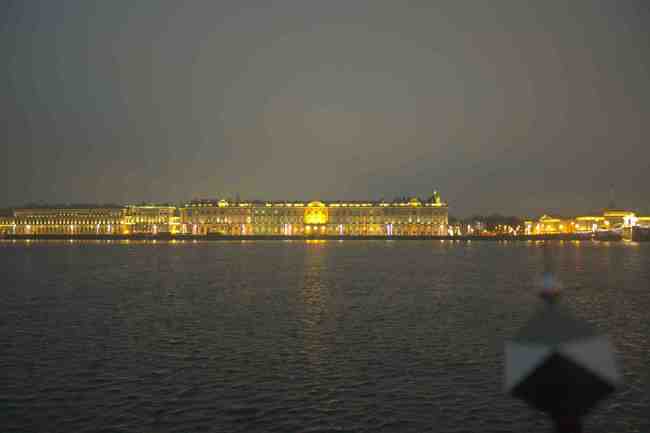}
    \label{fig:has_other_2}}
    \hfill
    \subfloat[]{
    \includegraphics[width=0.48\linewidth]{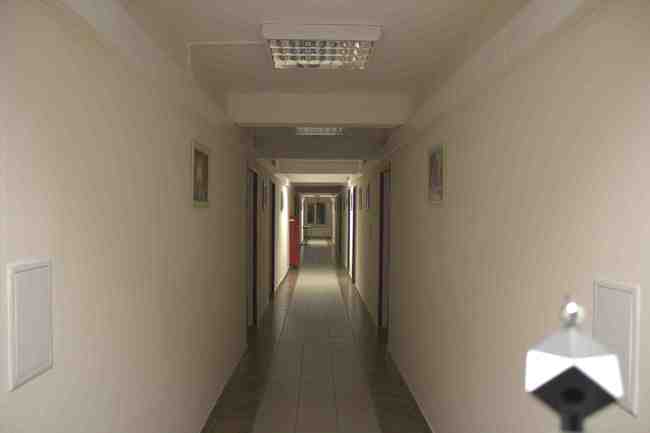} % c) Egor agreed, that INTERREFLECTION are relevant (Alex)
    \label{fig:cube_other}
    }
    % \hfill
    \subfloat[]{
    \includegraphics[width=0.48\linewidth]{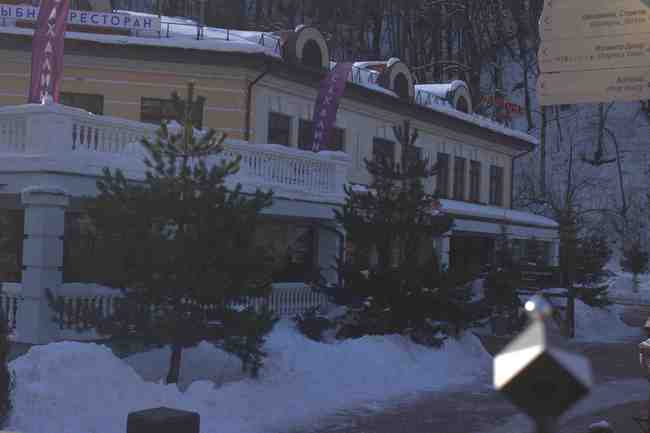}
    \label{fig:interreflection}
    }
    \end{center}
    \caption{Examples of images with partial illumination estimation: a) and~b) some scene parts are illuminated by the light source not captured by the cube; c) illumination significantly varies in the scene due to the interreflections d) all of the scene is in the shadow, while one of the cube faces is illuminated by the sun.}
    
    \label{data_collection:partial_examples}
\end{figure}

During the final quality filtration, all images were divided into three categories: a) images with \textbf{full light source estimation}, where the cube was illuminated by all the main light sources in the scene, b)~\textbf{incorrect} images where the cube does not allow to determine the illumination rating consistent with the scene, and c) the rest i.e. images with \textbf{partial light source estimation}, see Fig.~\ref{data_collection:partial_examples}.

As a result, about 400 images were marked as incorrect and removed from the dataset, about 524 images were marked as difficult images with a partial light source estimation, and the rest of the images were marked as good.

% Additionally, some photos were ruined by the fiber on the top of the cube, because sometimes it fell on the cube in the photo and this did not allow a successful ground-truth extraction. 
Additionally, the fiber on the top of the cube may fall on cube or remain on image after cropping out the color target. To prevent it, the fibers were glued to the cube or just cut off on most images. 
All of the images are captured horizontally without the use of camera flash.

\subsection{Ground-truth extraction}
\label{subsec:gt_estimation}

The ground-truth extraction was performed on raw images. First, a simple debayering has been performed by transforming each RGGB Bayer pattern square into a single pixel. The red and blue channels of the pixel color were obtained directly from the R and B components of the pattern, while the green was obtained by averaging the two G values. No interpolation was performed and therefore the number of image rows and columns was halved. 
Next, the oversaturated pixels were masked out and then the black level of $2^{11}$ was subtracted from all pixels. 
Finally, the ground-truth illumination values were extracted by calculating the average chromaticity of the manually annotated areas of the SpyderCube triangles. 

Four chromaticities were calculated for every image. 
They correspond to white and gray triangles on the left and right cube faces.
Note that on the brightly illuminated cubes, the white triangles may have oversaturated areas that cannot be properly used. 
On the contrary, the gray triangles chromaticities on the darker images may not be stable due to the black level noise.
It is important to note no image contains saturated grey edges, while some of the images contain saturated white edges and in such cases, a corresponding mark is provided.

The illuminations for a triangle were calculated as the mean illumination of its area after 50\% downscale to the barycenter.
%50\% downscale is 25\% area
The value of 50\% is selected as a simple empirical trade-off. 
Namely, a full-size triangle may contain non-triangle pixels because of unfocused cube or markup inaccuracies, while a tiny triangle would contain too few pixels and would be affected by noise. 

\subsection{Semantic markup}
\label{subsec:markup}

When developing and testing an algorithm for illumination estimation in a scene, it is useful to be able to analyze the structure of errors. The average error over the entire dataset will often not help to reveal whether e.g. the accuracy of the method for indoor images is much less accurate than for outdoor images. To enable performing such and similar checks faster and easier, additional information about the scene and shooting conditions were added to each image in the dataset. In addition to the information available during the shooting, this also includes following manual annotation:

\begin{itemize}
    \item Time of day (field \textbf{daytime}, with values day / night / unknown).
    \item The presence of objects with known coloration (\textbf{has\_known\_objects} field with values true / false).
    \item Scene illumination type (\textbf{illumination} field with values artificial / natural / unknown). It is worth noting that there are no flash photos in the dataset.
    \item Image sharpness (\textbf{is\_sharp} field with values true / false).
    \item The presence of light sources in the scene (the \textbf{light\_objects} multiple choice field with the values lamp / sky / sun / none). 
    \item The place where the image was captured (the \textbf{place} field with the values outdoor / indoor / unknown).
    \item Scene richness (field \textbf{richness} with values rich / simple).
    \item The presence of shadows in the scene (\textbf{shadows} field with values yes / no / unknown)
    \item The cube illumination by all the main light sources in the scene (\textbf{estimation} field with values full / partial)
\end{itemize}

Finally, it is important to note that none of the fields had a preset default value. In that way, the value of every field had to be explicitly set by an annotating person. 
Namely, if some default field values were to be set in advance, it could increase the annotation bias.

%% file: chapters/5_dataset.tex
\section{The proposed dataset}
\label{sec:proposed}

Having in mind all of the concerns and motivation from the previous section, a new dataset named Cube++ is proposed that continues on the previous Cube+ dataset. 
The dataset download link, the accompanying code, and the technical file description are available at \url{https://github.com/Visillect/CubePlusPlus/}. 

The Cube++ dataset contains $4890$ images. It includes only $1359$ of the $1707$ images from the Cube+ dataset and only $330$ of the $363$ images from the 1st Illumination Estimation Challenge (IEC2019) test set~\cite{iec2019}. 
Other images were excluded because they may go against respecting privacy by containing personal data such as faces and license plates or they may be problematic for ground-truth extraction. 
The remaining $3201$ images are brand new.

\begin{figure}[ht]
    \centering
    \includegraphics[width=\linewidth]{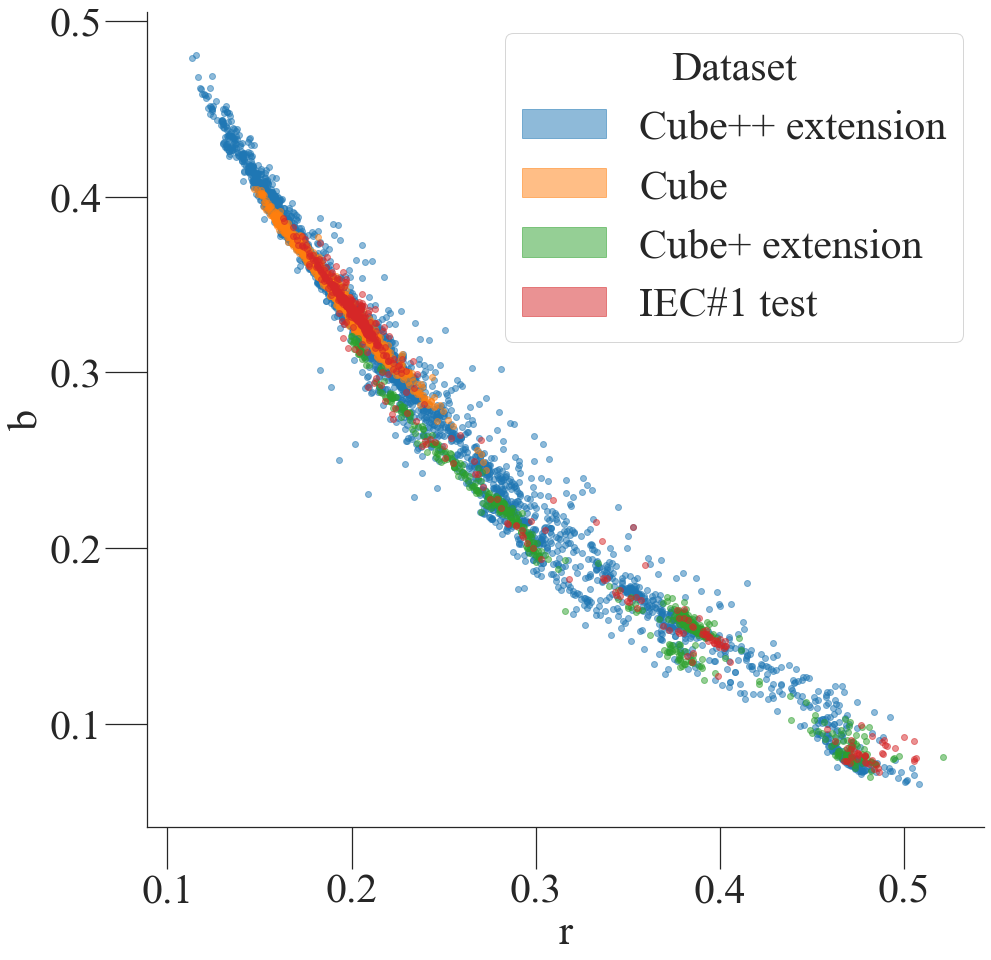}
    \caption{Scatter plot of illumination chromaticities captured by the SpyderCube gray faces for Cube++ and its parts.}
    \label{fig:2dscatterplot}
\end{figure}

Cube++ has diverse scene illumination cases as demonstrated by Fig~\ref{fig:2dscatterplot}.
There it can be seen that the chromaticity coverage area is wider than in e.g. Cube+.
In other words, the illumination variability has been significantly improved.

\begin{figure}[ht]
    \centering
    \includegraphics[width=\linewidth]{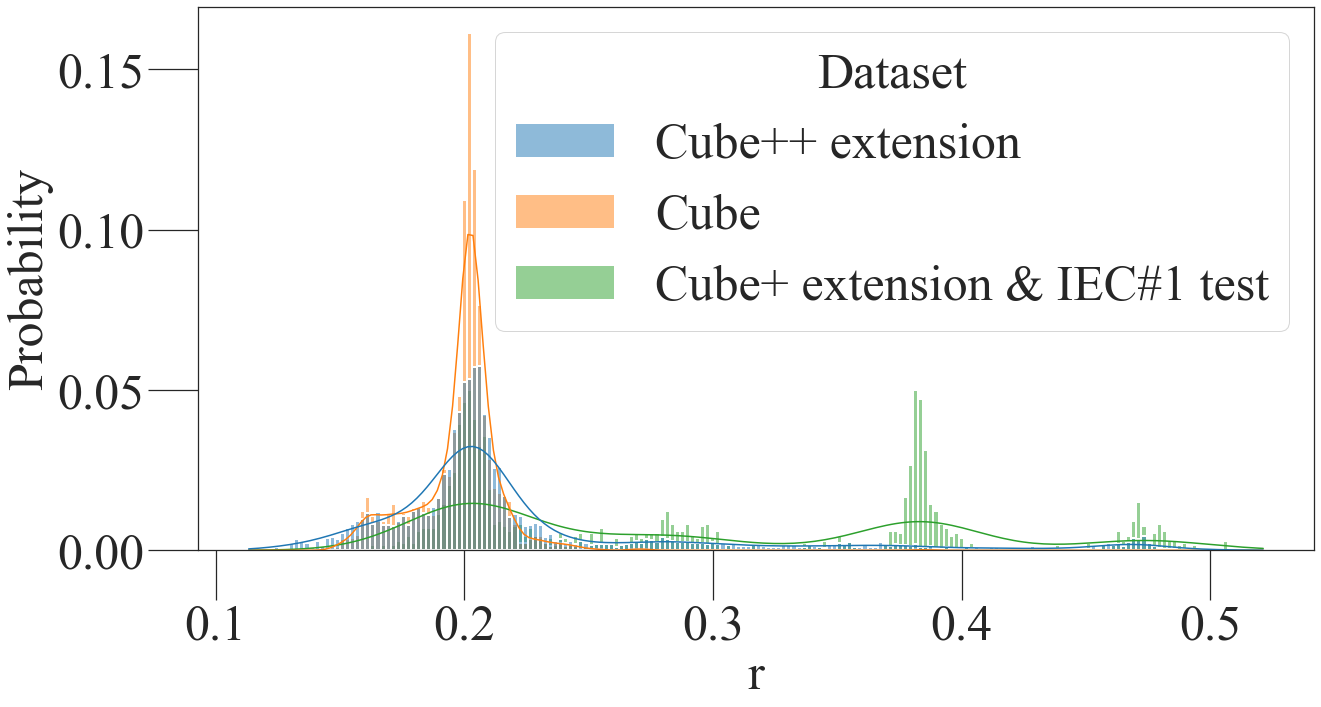}
    \caption{Histogram of the red chromaticity values $r=R/(R+G+B)$ for Cube++ ground-truth illuminations.}
    \label{fig:1d_histogram}
\end{figure}

The ground-truth illumination distribution can also be seen from another point of view by taking a look at Fig.~\ref{fig:1d_histogram}.
This figure shows that Cube+ and Cube++ have similar distributions, which in turn means that a lot of images were taken under outdoor daylight illumination.

\begin{figure}[ht]
    \centering
    \includegraphics[width=\linewidth]{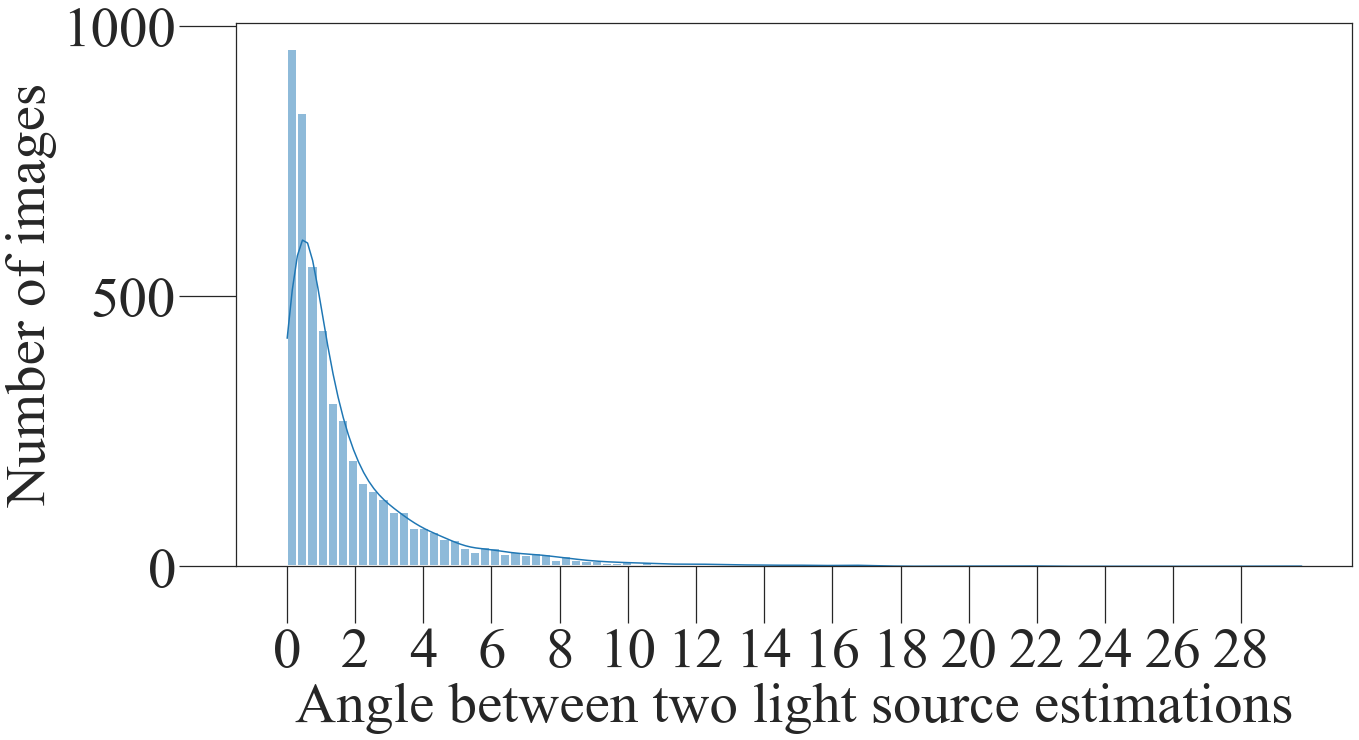}
    \caption{Histogram of angular differences between SpyderCube's left and right gray faces for Cube++.}
    \label{fig:angle_difference}
\end{figure}

One of the important features of the proposed dataset is the fact that it contains two ground-truth illumination records per image, one for each side of the SpyderCube instance. 
Even though in many of the images there is effectively only one dominant illumination in the scene, Fig.~\ref{fig:angle_difference} helps to better understand the relation between the two recorded illuminations over the dataset images. 
Currently the average angular error of state-of-the-art illumination estimation methods is arguably somewhere between $1^\circ$ and $2^\circ$. 
With that in mind, all images with larger angular difference between their illumination records can be treated as two-illumination cases.

Another important feature of the proposed dataset that has to be stressed additionally is that it contains semantic data for each image. 
All semantic information features are shown in Table~\ref{tab:datasetsfeatures}. 
Different features provided in the semantic data can be helpful for algorithm tuning and profiling as they give potentially useful information about each image individually.
% Different image features can be used for algorithms tuning and profiling.

\subsection{Technical description}
\label{subsec:technical}

The dataset consists of several parts. 
First, there are the raw images with only simple linear debayering performed that are stored in 16-bit PNG format. 
Next, there are CSV files with ground-truth illumination values and CSV files with additional related properties. 
Furthermore, there are JPEG images generated by using the \texttt{dcraw} open-source tool\footnote{\url{https://www.dechifro.org/dcraw/}}. 
Finally, there are also additional files for storing auxiliary information. 
All these files are automatically built from sources included in the dataset by running a script that is also provided. 
The sources contain the original CR2 images from the camera and JSON files with the manual annotation data. 

The original camera JPEG images are not included as their generation depends on cameras' settings, which means that they cannot be recreated simply or even accurately~\cite{kim2012new}.

\subsubsection{PNG and JPEG images}
\label{subsubsec:png_jpeg}

The main 16-bit PNG images are generated from the original CR2 files in three steps.
First, the CR2 files are decoded by using the \texttt{dcraw} tool with the options \texttt{-D -4 -T}. 
This generates a 16-bit 1-channel TIFF image.
Second, the $[10, 10+5184] \times [4,4+3456]$ rectangle was cropped, to have the same area as the default camera JPEG, which comes with certain advantages. Finally, a naive debayering is applied so that every $\begin{smallmatrix}R,&G_1,\\G_2,&B\end{smallmatrix}$ pattern is converted to a pixel of color $\left(R, \frac{G_1 + G_2}{2}, B\right)$. 
After that the size of the generated PNG images is $2592 \times 1728 = 2^5 3^4 \times  2^6 3^3$. Even though the color channel values have 16 bits of storage, in practice their maximal value is always below $2^{14}-1$. The black level that can be used for every dataset image is $2048=2^{11}$. 

For visualization purposes, the modified versions of JPEG images generated by the \texttt{dcraw} tool are included as well.
The modification includes cropping and downscaling in order for the JPEG images to have the same size as the PNG images. Downscaling is required because JPEG images generated by \texttt{dcraw} are not downscaled like the PNG images. 
On the other hand, JPEG images generated by the camera were not included because they depend on camera settings and the camera's white balancing algorithm, which is proprietary, not fully documented, and it may differ for Canon 550D and 600D cameras that have been used for image capturing. Because of that, they can not be recreated reliably.

\subsubsection{The ground-truth}
\label{subsubsec:gt}

The ground-truth illumination records are stored in the \texttt{gt.csv} file.
Ground-truth illuminations are calculated as described in Section~\ref{subsec:gt_estimation}.
The columns are: \textit{image} and for each of the 4 triangles (left, right, left bottom, right bottom) it contains three columns \textit{r, g, b} with the corresponding RGB illumination chromaticities so that $r + g + b = 1$. 
The triangles brightness values are given in the \texttt{properties.csv} file.

Usually, computational color constancy datasets contain only a single ground-truth illumination vector, which represents the dominant illumination in the scene. 
In the Cube++ dataset such illumination is not given, because the precise single illuminant estimation may require the specialist annotation. 
Moreover, some images have two significantly different illuminations, which makes it harder to select the dominant one. 
If only a  single ground-truth illumination is required and the possible errors that it leads to are acceptable, then one of the following methods can be used to obtain it: \begin{itemize}
    \item sample images with relatively similar left and right ground-truth illuminations (the sugested answers for such images are denoted in \textit{properties.csv});
    \item select from the left and right sides the brighter one; then select the white triangle for a dark image, and grey triangle for a bright image.
\end{itemize}

Note that the difference between the sensitivities of the white faces is greater then the difference between the sensitivities of the gray ones (see Section~\ref{subsec:setup}). 
Additionally, since the white faces are more often overexposed than the gray ones, using the gray faces should be preferred. 
On the contrary, using white faces may be better on dark images as mentioned in Section~\ref{subsec:gt_estimation}.

We also estimated if the ground truth values are distorted by the pixels with the clipped values. The images with overxposed grey triangles were removed from the dataset. The images with the clipped values on white triangles are present in the dataset, but the overexposed triangles are marked in \textit{properties.csv}.

\subsubsection{Relevant meta-information}
\label{subsubsec:meta}

The \texttt{properties.csv} file contains the most relevant meta-information about the Cube++ images. It includes the average triangle brightness $\frac{R+B+G}{3}$, manual annotation data, information about overexposed triangles, and a carefully selected subset of EXIF data fields.

The EXIF data was extracted from CR2 files using the \texttt{PyExifTool} library\footnote{\url{https://pypi.org/project/PyExifTool/}}.
All the extracted values can be found in the corresponding JSON files. The properties table contains only a few selected ones. The EXIF data format slightly differs between the Canon EOS 550D and 600D cameras: there are 312 common fields, 2 in 550D only, and 21 in 600D only. All the selected EXIF fields are common. 

The \texttt{cam\_estimation.csv} file contains the EXIF fields of the camera that contains the camera's light source estimation

\newcommand\turnds[1]{#1}
\newcommand\twoword[2]{\begin{tabular}[c]{@{}c@{}} {#1} \\ {#2} \end{tabular}}
\addtocounter{footnote}{1}
\newcommand\fm{\footnotemark[\value{footnote}]}

\begin{table*}[ht]
\normalsize
\caption{Feature statistics for various Cube++ subsets}
\label{tab:datasetsfeatures}
\centering
\begin{tabular}{|p{3cm}|c|c|c|c|c|c|c|c|}
    \hline
    \multicolumn{2}{|c|}{{\multirow{2}{*}{\textbf{Feature}}}} & \multicolumn{5}{c|}{{Cube++ parts}} &  \multicolumn{2}{c|}{{Cube++ subsets}} \\ \cline{3-9}
    \multicolumn{2}{|c|}{} & {\turnds{Cube\fm}} & {\turnds{\twoword{Cube+\fm}{extension}}} & {\turnds{\twoword{ {IEC2019\fm} }{test}}} &  {\turnds{\twoword{Cube++}{extension}}} & \textbf{\turnds{\twoword{Total}{Cube++}}} & {\turnds{\twoword {Full}{estimation}}} & \textbf{\turnds{\twoword{Simple}{Cube++}}} \\
\hline\multirow{2}{*}{\textbf{estimation}}
                                     & full       & 916  & 306   & 301        & 2843   & \textbf{4366}  & 4366 & \textbf{2234}             \\
                                     & partial    & 115  & 22    & 29         & 358    & \textbf{524 }  & 0    & \textbf{0   }             \\ 
\hline\multirow{3}{*}{\textbf{daytime}}
                                     & day        & 1007 & 24    & 176        & 2302   & \textbf{3509}  & 3155 & \textbf{1615}             \\
                                     & night      & 1    & 1     & 1          & 71     & \textbf{74  }  & 36   & \textbf{5   }             \\
                                     & unknown    & 23   & 303   & 153        & 828    & \textbf{1307}  & 1175 & \textbf{614 }             \\ 
\hline\multirow{2}{*}{\textbf{has\_known\_objects}}
                                     & True       & 767  & 49    & 76         & 1555   & \textbf{2447}  & 2185 & \textbf{1138}             \\
                                     & False      & 264  & 279   & 254        & 1646   & \textbf{2443}  & 2181 & \textbf{1096}             \\ 
\hline\multirow{4}{*}{\textbf{illumination}}
                                     & artificial & 2    & 9     & 16         & 306    & \textbf{333 }  & 233  & \textbf{109 }             \\
                                     & mixed      & 0    & 0     & 1          & 27     & \textbf{28  }  & 13   & \textbf{7   }             \\
                                     & natural    & 995  & 41    & 164        & 2311   & \textbf{3511}  & 3171 & \textbf{1611}             \\
                                     & unknown    & 34   & 278   & 149        & 557    & \textbf{1018}  & 949  & \textbf{507 }             \\ 
\hline\multirow{2}{*}{\textbf{is\_sharp}}
                                     & True       & 1002 & 259   & 309        & 3006   & \textbf{4576}  & 4088 & \textbf{2088}             \\
                                     & False      & 29   & 69    & 21         & 195    & \textbf{314 }  & 278  & \textbf{146 }             \\ 
\hline\multirow{7}{*}{\textbf{light\_objects}}
                                     & none       & 560  & 326   & 287        & 1505   & \textbf{2678}  & 2434 & \textbf{1246}             \\
                                     & flash      & 0    & 0     & 0          & 3      & \textbf{3   }  & 3    & \textbf{2   }             \\
                                     & lamp       & 1    & 0     & 1          & 125    & \textbf{127 }  & 58   & \textbf{29  }             \\
                                     & lamp\&sky  & 0    & 0     & 0          & 21     & \textbf{21  }  & 7    & \textbf{5   }             \\
                                     & sky        & 470  & 2     & 42         & 1543   & \textbf{2057}  & 1861 & \textbf{952 }             \\
                                     & sky\&sun   & 0    & 0     & 0          & 3      & \textbf{3   }  & 3    & \textbf{0   }             \\
                                     & sun        & 0    & 0     & 0          & 1      & \textbf{1   }  & 0    & \textbf{0   }             \\ 
\hline\multirow{3}{*}{\textbf{place}}
                                     & indoor     & 12   & 281   & 27         & 454    & \textbf{774 }  & 653  & \textbf{379 }             \\
                                     & outdoor    & 1017 & 46    & 217        & 2536   & \textbf{3816}  & 3432 & \textbf{1738}             \\
                                     & unknown    & 2    & 1     & 86         & 211    & \textbf{300 }  & 281  & \textbf{117 }             \\ 
\hline\multirow{3}{*}{\textbf{richness}}
                                     & rich       & 932  & 152   & 233        & 2840   & \textbf{4157}  & 400  & \textbf{198 }             \\
                                     & simple     & 3    & 8     & 94         & 337    & \textbf{442 }  & 3704 & \textbf{1860}             \\
                                     & unknown    & 96   & 168   & 3          & 24     & \textbf{291 }  & 262  & \textbf{176 }             \\
\hline\multirow{3}{*}{\textbf{shadows}}
                                     & no         & 767  & 125   & 259        & 1879   & \textbf{3030}  & 1573 & \textbf{696 }             \\
                                     & yes        & 264  & 203   & 61         & 1282   & \textbf{1810}  & 2751 & \textbf{1519}             \\
                                     & unknown    & 0    & 0     & 10         & 40     & \textbf{50  }  & 42   & \textbf{19  }             \\ 
\hline\textbf{Total}                               
                                     &            & 1031 & 328   & 330        & 3201   & \textbf{4890}  & 4366 & \textbf{2234}             \\ 
\hline
\end{tabular}
\end{table*}

\footnotetext{Not including Cube, Cube+, IEC2019test images, removed from the Cube++ dataset because of GDPR restrictions or possible problems with ground-truth extraction.}

\subsection{Image preparation}
\label{subsec:preparation}

Finally, it is important to clearly specify how to properly prepare the provided Cube++ images before handing them over to illumination estimation methods that are to be tested.

There are three main steps that have to be taken. 
\subsubsection{Black level substraction.} The first step is to subtract the approximate black level of $2048$ from all image pixel color components. 
In some cases this can result in negative values, but such values should then be set to $0$.

\subsubsection{Saturation detection.} The second step is to calculate the maximum value $m$ for all pixels across all color channels. After that all pixels that have a value greater than or equal to $m-50$ in any of their channels should have all their channel values set to $0$
This would remove most of the incorrect pixels with clipped values. 
Nethertheless, it would leave some rare overexposed pixels, because demosaicing procedure may mix them with the normal ones.
To get precise information about saturated pixels it is recommended to analyse images before demosaicing (the last one can be extracted from CR2 files).

\subsubsection{Color target masking.} The last step is to mask out the lower right rectangle of the image that contains the color target to remove any potential bias and thus to have a relatively fair testing. The size of this rectangle is $700\times 1000$ for all images. The rectangle is masked out by setting all channel values of all its pixels to $0$, i.e. by making it black.

%The execution of all these steps is demonstrated by the script that is publicly available at~\url{https://github.com/Visillect/CubePlusPlus/}.

% contains all the images data, but there is a lot of unrelated technical details. It contains files: 
% \begin{itemize}
%     \item list.csv - file with information of subdatasets, where the image is present
%     \item CR2 - directory with raw CR2 images
%     \item PNG - directory with raw PNG images
% The smaller dataset contains downscaled images with resolution $1296\times864 = 3\cdot2^4 3^3  \times 2 \cdot2^4 3^3$ and can be downloaded easily.
% 

% The dataset includes the images of the Cube+ dataset and the training images of the 2nd Illumination Estimation Challenge.

% \todo[inline]{two datasets version}
% The dataset consists of two parts: the main and the smaller one.
% The main one contains all the images data, but there is a lot of unrelated technical details. It contains files: 
% \begin{itemize}
%     \item list.csv - file with information of subdatasets, where the image is present
%     \item CR2 - directory with raw CR2 images
%     \item PNG - directory with raw PNG images
    
% \end{itemize}

\begin{figure}[htb]
    \centering
    
	\subfloat[]{
	\includegraphics[width=0.48\linewidth]{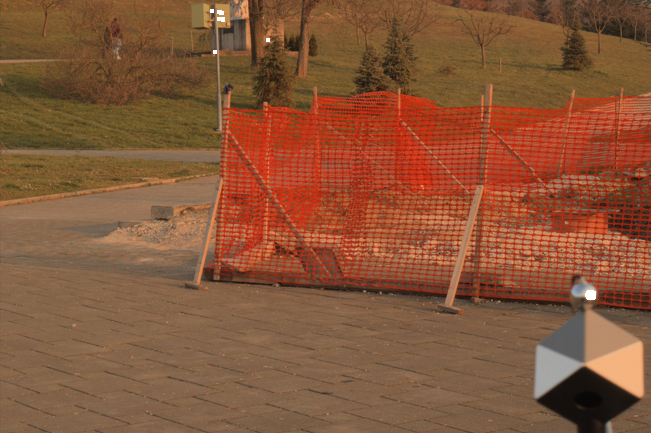}
	\label{fig:left}
	}%
	~%
	\subfloat[]{
	\includegraphics[width=0.48\linewidth]{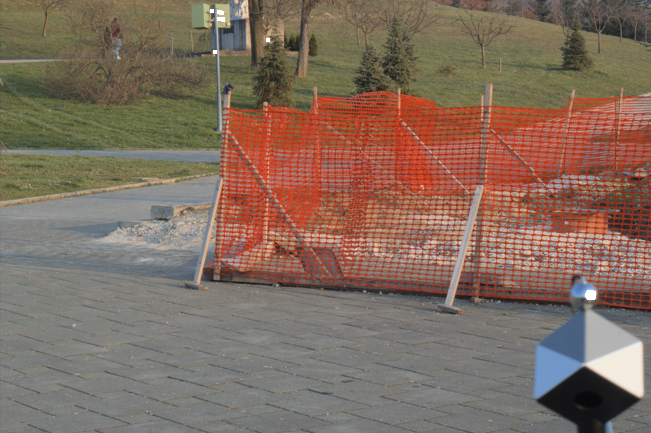}
	\label{fig:right}
	}
	
    \caption{Example of chromatic adaptation based on illumination extracted from a) left and b) right gray face of the SpyderCube calibration object placed in the scene.}
	\label{fig:two}
    
\end{figure}

\subsection{Intended dataset usage}
\label{subsec:usage}

With all its features, especially the two ground-truth illumination records, Cube++ is appropriate for several illumination estimation use cases.
All datasets mentioned in Section \ref{sec:existing}, except for maybe TCC dataset, are designed for the most widely used classical illumination estimation problem: estimation of the single light source in the scene.
Therefore, each image is provided with only single light source ground-truth, even in cases when the scene is obviously under the influence of multiple illuminations. In contrast to these datasets, Cube++ allows to work on following problem statements:
\begin{enumerate}
    \item Estimate one and only dominant lighting in a scene;
    \item Estimate two dominant light sources in a scene;
    \item Estimate at least one dominant light source in the scene.
\end{enumerate}

For each of the listed problem statements we propose the following rules to filter the Cube++ images that are suitable for it.
To form the dataset subset for the first problem, one needs to select all images where the angular differences between its two extracted ground-truth illuminations is below $1^\circ$ (except partially light source estimation part, see section \ref{subsec:collection_process}). 
For the two light source estimation problem, one needs to do the opposite, i.e. to select all images that are not selected for the first problem (except partially light source estimation part).
Finally, to work on the third problem, all images can be chosen.

Here it is important to mention that the proposed rules are arbitrary, that they will result in some of the images being inappropriately selected, and that they may be improved. One example where these rules fail is shown in Fig.~\ref{fig:two}. There the extracted ground-truth illuminations differ significantly, but practically the whole scene is mostly under the illumination captured by the right cube face. 
This means that even though the scene is effectively under uniform illumination, the mentioned rules will result in the opposite conclusion based on the difference between the extracted illuminations on the two cube faces.

Also it is important to mention what at the moment partially light source estimation part of the dataset is not provided with subjective single illumination estimation choice.
The plan is to solve this in future work.

\subsection{SimpleCube++ dataset}
\label{subsec:small_ds}

In addition to the main 200GB Cube++ dataset, a 2GB-small and simpler version of it is prepared. 
The small dataset contains 4x downscaled images which have less than $1^\circ$ degrees difference between left and right grey edges ground truth illumination estimation. It includes only images with single source illumination, consequently ground truth file contains only one ground truth per image. This one ground truth was extracted in a following manner: firstly, average values for both gray triangles were calculated as in the main Cube++ dataset; secondly, they were normed with $l_{1}$-norm ($r+g+b=1$ for both gray triangles); finally, obtained ground truth values were averaged and normed again with $l_{1}$-norm. 
This dataset has two main advantages: small weight (around 2GB) and single answer. 

SimpleCube++ contains PNG and JPG files, \texttt{gt.csv} with ground truth data and \texttt{properties.csv} with manual annotation data. 
In addition, small dataset was divided into train and test parts with rule: each image was independently assigned to the test set with probability $20\%$.

%% file: chapters/6_discussion.tex
\section{Discussion}
\label{sec:discussion}

Having another high-quality illumination estimation dataset such as the one proposed in this paper is certainly beneficial to the interested research community as well as the industrial sector and there should probably be no discussion about that.
However, proposing a new dataset is still only an incremental move in terms of the overall paradigm of illumination estimation research since this has been done on numerous occasions while the dataset usage has remained relatively unchanged.

A much more constructive and necessary discussion that is rarely taken forward should be about the direction of how to better use or not use the datasets to achieve better progress in illumination estimation research. 
In terms of that, one of the burning issues is that the results in most illumination estimation papers are unverifiable and thus questionable. Therefore, for the sake of improving the state of the illumination estimation research, it would be quite useful to further discuss this problem as well as the potential solutions to it in more detail.

\subsection{Questionable research progress}
\label{subsec:progress}

Obtaining low illumination estimation errors on a benchmark dataset is a regularly used approach when trying to demonstrate the superiority of a proposed illumination estimation method. 
For all well-known datasets the ground-truth illumination used during the test phase is publicly available and the actual error statistics calculation is usually performed by the authors themselves and published in their papers. However, this introduces several problems with the most serious being data dredging, i.e. $p$-hacking and erroneous reporting.

The problem with data dredging in illumination estimation is that in cases when a model selection is required, the final results that are reported were not always obtained through nested cross-validation~\cite{deisenroth2020mathematics}.
Instead, the reported results are the ones that were used to select the model in the first place. 
By using these results, a method's true performance on new unknown data may be masked and unfairly shown to be better than it actually is. 
This can prevent or slow down the progress in illumination estimation research by giving misleading clues about the validity of the method's underlying assumptions.

In the area of visual odometry similar problems with e.g. the KITTI dataset~\cite{geiger2012we} have been prevented by simply keeping the ground-truth for the test secret. 
By having the evaluation of the results on the test set carried out by the dataset administrators, any serious attempts of $p$-hacking have been prevented.

Another problem that can be prevented if the evaluation is carried out by a third party is the erroneous reporting. 
For example, in~\cite{cheng2014illuminant} the results of the proposed illumination estimation method on several datasets were allegedly all obtained by using the same value of a hyperparameter. 
However, trying to re-implement the method fails to produce the same results and only after checking the associated webpage~\cite{nus2020online} it becomes clear that the hyperparameter value has to be changed for each dataset to fully reproduce the published results.

A somewhat similar example is the 2007 paper by van de Weijer et al.~\cite{van2007using}. 
In an erratum published in 2008~\cite{van2020online} it was explained how testing was inadequately performed, which consequently resulted in reporting of erroneous error statistics.

Finally, any doubts in the validity of some reported error statistics could be reduced or fully eliminated if they were calculated not by the authors themselves, but by a reliable third party. 
This would also help the overall research progress.

\subsection{Illumination estimation challenges}
\label{subsec:challenges}

Inspired by the ideas mentioned in the previous subsection, two international illumination estimation challenges have already taken place~\cite{iec2019,iec2020}. The challenges provided the participants with thousands of training images and their respective ground-truth illuminations, while for the test set only the images were provided and the ground-truth remained secret until the end of the challenge. Because of that the error statistics for the illumination estimations sent over by the authors were calculated by the challenge organizers, which prevented a lot of problems described in the previous section. The results were thus more trustworthy and they have shown e.g. high errors for some methods that were previously reported to be highly accurate. Additionally, the challenges helped to recognize additional problems such as training a method to obtain excellent values for a given error metric~\cite{savchik2019color}, which results in issues related to the so called Goodhart's law~\cite{strathern1997improving}.

\subsection{Benchmark}

While the described international illumination estimation challenges have shown the advantages of having a reliable third party calculate the error statistics, they were fixed in time and they cannot be repeated on the same images anymore. Therefore, the next step would be to create a benchmark dataset similar to the KITTI dataset with an online user interface for submitting the results at any given time. This would surely represent a significant contribution to the illumination estimation research since it would simultaneously provide the researchers with trustworthy results and also eliminate many of the serious problems that were described earlier in this paper.

For the above reasons, creating such a benchmark is already underway at the time of writing this paper. 
Present time we are working on the question of benchmark creation.
Possible benchmark will be based on the images that were taken during the same time as the rest of the Cube++ images, but that were excluded from its final version. 
Because of that, in this paper there are purposely no error statistics obtained on the Cube++ dataset by any of the illumination estimation methods. The error statistics will be published online and they will be based on the first version of the benchmark test set. This aims to avoid providing any results obtained on the Cube++ images with known ground-truth illumination. Namely, the idea is to separate the testing and the associated problems from the dataset and to relegate it to the benchmark. 
Therefore, the overall goal of this paper is to provide high quality training data without any testing. The role of testing data is to be assumed by the future benchmark.

%% file: chapters/7_conclusion.tex
\section{Conclusions}
\label{sec:conclusions}

A new illumination estimation dataset named Cube++ has been proposed. 
Unlike similar existing illumination estimation datasets, it provides rich, reliable, and verifiable data on scene illumination with specific care being given to precise calibration. 
For every of its 4890 images there are two ground-truth illumination records as well as a multitude of semantic information and it is GDPR-compliant. 
Furthermore, a wide variety of scene content is covered, and numerous illuminations are captured. 
Cube++ contains images taken with several instances of the same model of the camera sensor.
In addition to that, a centralized versioning control system for Cube++ has been established to simplify and document possible future changes in the dataset and error handling. 
By having these properties and novelties, Cube++ is technically superior to most similar illumination estimation datasets. 
One of the future steps that should also be a significant progress in the overall illumination estimation research is to create an online illumination estimation benchmark based on the infrastructure that was used to create the Cube++ dataset.

%% file: chapters/8_acknowledgements.tex
\section*{Acknowledgment}
\label{sec:acknowledgment}
The authors would like to thank Prof. Dmitry Nikolaev, for his help in reading the paper and his contribution to the discussions.

The authors would also like to thank Maria Yarykina and Ekaterina Panfilova for the images they added to the dataset and Viacheslav Vasilyev, Vasily Tesalin, Sergey Emelyanov, Oleg Emelyanov, Tatyana Postnikova, Sergey Pavlov, Evgenija Sidorchuk, and Olga Vlasova for their contribution to the annotation.

%% file: chapters/bio.tex
\begin{IEEEbiography}[{\includegraphics[width=1in,height=1.25in,clip,keepaspectratio]{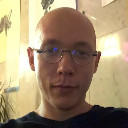}}]{Egor Ershov} 
(ORCID=\url{https://orcid.org/0000-0001-6797-6284})
received a Ph.D. degree from the Faculty of Radio Engineering and Computer Technology in 2019. He is a senior researcher in Vision System Laboratory in Institute of Information Transmission Problems (Kharkevic Institute) of the Russian Academy of science (IITP RAS).
He is also senior lecturer at Computer Science Faculty in Higher School of Economy.
His main areas of research are image processing and analysis, particularly color computer vision, color reproduction technologies, colorimetry, human vision system, Hough transform.
He is an assistance manager of the cathedra at Computer Science faculty in Higher School of Economy.
He is a recipient of several awards for his scientific and professional work.

\end{IEEEbiography}

\begin{IEEEbiography}[{\includegraphics[width=1in,height=1.25in,clip,keepaspectratio]{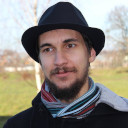}}]{Alexey (Alex) Savchik}
(ORCID=\url{https://orcid.org/0000-0003-3035-1365})
Vision  Systems  Laboratory,  Institute   for   Information   Transmission   Problems (Kharkevich Institute) of the Russian Academy of Sciences (IITP RAS), Moscow, Russia. Alex Savchik in 2014 graduated from Department of Mechanics and Mathematics of Moscow State Univercity, in 2012 from Yandex School of Data Analysis. His research interests include deep learning and computer vision.

\end{IEEEbiography}

\begin{IEEEbiography}[{\includegraphics[width=1in,height=1.25in,clip,keepaspectratio]{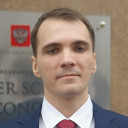}}]{Illya (Ilya) Semenkov} (ORCID=\url{https://orcid.org/0000-0003-1515-7062}) Vision  Systems  Laboratory,  Institute   for   Information   Transmission   Problems (Kharkevich Institute) of the Russian Academy of Sciences (IITP RAS), Moscow, Russia. Received Bachelor's degree at National Research University Higher School of Economics (NRU HSE) in 2019 and is currently Master's student at Faculty of Computer Science of NRU HSE. His research interests include Deep Learning, Computer Vision, Statistical Learning and Optimal Transport. 
\end{IEEEbiography}

\begin{IEEEbiography}[{\includegraphics[width=1in,height=1.25in,clip,keepaspectratio]{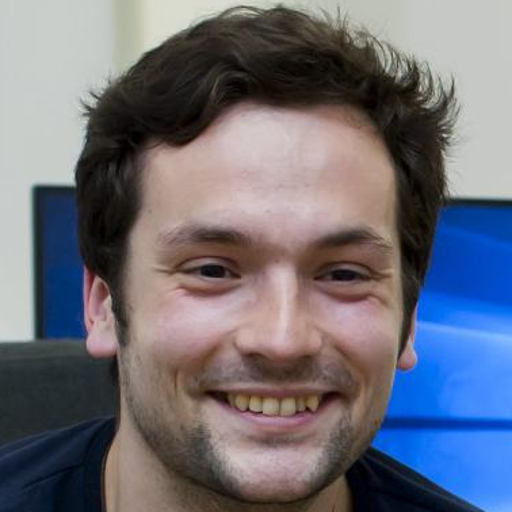}}]{Nikola Bani{\'{c}}} (ORCID=\url{https://orcid.org/0000-0002-3900-8590}) received B.Sc., M.Sc., and Ph.D. degrees in computer science in 2011, 2013, and 2016, respectively. He is currently working as a senior computer vision engineer at Gideon Brothers, Croatia. He has worked in real-time image enhancement for embedded systems, digital signature recognition, people tracking and counting, and image processing for stereo vision. His research interests include image enhancement, color constancy, image processing for stereo vision, and tone mapping.
\end{IEEEbiography}

\begin{IEEEbiography}[{\includegraphics[width=1in,height=1.25in,clip,keepaspectratio]{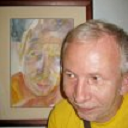}}]{Alexander Belokopytov} 
(ORCID=\url{https://orcid.org/0000-0001-9698-7206})  Vision Systems Laboratory, Institute for Information Transmission  Problems (Kharkevich Institute) of the Russian Academy of Sciences (IITP RAS), Moscow, Russia

Alexander Belokopytov in 1982 graduated from Moscow Institute of Physics and Technology (MIPT) with specialisation in microwave engineering. While in IITP, he participated in research in peripheral and color human vision, underwater photography. His research interests include human vision (color and peripheral), optoelectronics, and scientific data visualisation.
\end{IEEEbiography}

\begin{IEEEbiography}[{\includegraphics[width=1in,height=1.25in,clip,keepaspectratio]{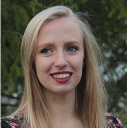}}]{Daria Senshina} (ORCID=\url{https://orcid.org/0000-0002-6287-4083}) Vision  Systems  Laboratory,  Institute   for   Information   Transmission   Problems (Kharkevich Institute) of the Russian Academy of Sciences (IITP RAS), Moscow, Russia. Received Bachelor's degree at Moscow Institute of Physics and Technology (MIPT) in 2020. Daria Senshina is currently Master's student at Phystech School of Applied Mathematics and Informatics of MIPT and simultaneously obtaining M.Sc. degree in Industrial and Applied Maths at the University Grenoble Alpes, Grenoble, France. Her research interests include computer vision and image processing.
\end{IEEEbiography}

\begin{IEEEbiography}[{\includegraphics[width=1in,height=1.25in,clip,keepaspectratio]{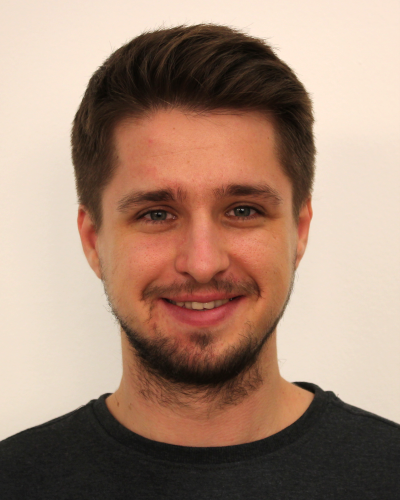}}]{Karlo Ko{\v{s}}{\v{c}}evi{\'{c}}} (ORCID=\url{https://orcid.org/0000-0002-9691-4231}) received B.Sc. and M.Sc. degrees in computer science in 2016 and 2018, respectively. He is currently in his second year of the technical sciences in the scientific field of computing Ph.D. program at the Faculty of Electrical Engineering and Computing, University of Zagreb, Croatia. His research interests include image processing, image analysis, and deep learning. His current research is in the area of color constancy with a focus on learning-based methods for illumination estimation.
\end{IEEEbiography}

\begin{IEEEbiography}[{\includegraphics[width=1in,height=1.25in,clip,keepaspectratio]{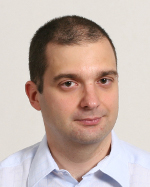}}]{Marko Suba{\v{s}}i{\'{c}}} (ORCID=\url{https://orcid.org/0000-0002-4321-4557}) received a Ph.D. degree from the Faculty of Electrical Engineering and Computing at the University of Zagreb in 2007. Since 1999, he has been working at the Department for Electronic Systems and Information Processing at the Faculty of Electrical Engineering and Computing at the University of Zagreb, currently as an associate professor. He teaches several courses at the graduate and undergraduate levels. His research interests lie in image processing and analysis and neural networks, with a particular interest in image segmentation, detection techniques, and deep learning. He is a member of the IEEE - Computer Society, the Croatian Center for Computer Vision, the Croatian Society for Biomedical Engineering and Medical Physics, and the Centre of Research Excellence for Data Science and Advanced Cooperative Systems. 
\end{IEEEbiography}

\begin{IEEEbiography}[{\includegraphics[width=1in,height=1.25in,clip,keepaspectratio]{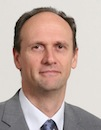}}]{Sven Lon{\v{c}}ari{\'{c}}}
(ORCID=\url{https://orcid.org/0000-0001-6797-6284})
received B.Sc., M.Sc., and Ph.D. degrees in 1985, 1989, and 1994, respectively. After earning his doctoral degree, he continued his academic career at the Faculty of Electrical Engineering and Computing, University of Zagreb, where he is currently a full professor. He was an assistant professor at the Department of Electrical and Computer Engineering, New Jersey Institute of Technology, NJ, USA, from 2001-2003. His main areas of research are image processing and analysis. Together with his students and collaborators, he has published more than 200 publications in scientific peer-reviewed journals and has presented his work at international conferences. He is a senior member of the IEEE, director of the Center for Artificial Intelligence, and co-director of the national Center of Research Excellence for Data Science and Advanced Cooperative Systems. He is a recipient of several awards for his scientific and professional work.
\end{IEEEbiography}